\documentclass[12pt]{article}  
\usepackage{multirow}
\usepackage{amssymb,graphicx,amsmath,color,algorithm,algorithmic}
\usepackage[numbers]{natbib}
\usepackage{caption}
\usepackage{subcaption}
\DeclareCaptionLabelSeparator{bar}{\space\textbf{\textbar}\space}
\DeclareCaptionFont{captionfont}{\fontsize{11pt}{11pt}\selectfont}
\usepackage{geometry}
\usepackage{url}
\usepackage[hidelinks]{hyperref}
\usepackage[open,openlevel=1]{bookmark}
\usepackage{array}
\usepackage{lineno}
\usepackage{tocloft}
\usepackage{parskip}
\usepackage{authblk}
\usepackage{soul,xcolor}
\usepackage{svg}
\usepackage{tikz}
\usepackage{xstring}

\newlength{\mylen}
\cftpagenumbersoff{figure}
\cftpagenumbersoff{table}

\makeatletter
\newcommand{\setword}[2]{%
  \phantomsection
  #1\def\@currentlabel{\unexpanded{#1}}\label{#2}%
}
\makeatother

\newcommand{\figcap}[1]{{\textbf{\MakeLowercase{#1}},}}
\DeclareMathOperator*{\argmax}{arg\max}

\title{\LARGE \bf Identifying Important Sensory Feedback for Learning Locomotion Skills
}
\author[1,*]{Wanming Yu}
\author[1,2,*]{Chuanyu Yang}
\author[1]{Christopher McGreavy}
\author[1]{Eleftherios Triantafyllidis}
\author[3]{Guillaume Bellegarda}
\author[3]{Milad Shafiee}
\author[3]{Auke Jan Ijspeert}
\author[4,$\dagger$]{Zhibin Li}

\affil[1]{University of Edinburgh, Edinburgh, UK.}
\affil[2]{Shenzhen Amigaga Technology Co Ltd, Shenzhen, China.}
\affil[3]{École Polytechnique Fédérale de Lausanne (EPFL), Lausanne, Switzerland.}
\affil[4]{University College London, London, UK.}
\affil[*]{These authors contributed equally: Wanming Yu, Chuanyu Yang.}
\affil[$\dagger$]{Corresponding author: alex.li@ucl.ac.uk}
\date{}

\usepackage{setspace}
\def \capA {a}
\def \capB {b}
\def \capC {c}
\def \capD {d}
\def \capE {e}
\def \capF {f}

\begin{document}
\maketitle
\vspace{-15mm}
\pdfbookmark[1]{Abstract}{Abstract}
\section*{Abstract}
Robot motor skills can be learned through deep reinforcement learning (DRL) by neural networks as state-action mappings. While the selection of state observations is crucial, there has been a lack of quantitative analysis to date. Here, we present a systematic saliency analysis that quantitatively evaluates the relative importance of different feedback states for motor skills learned through DRL. Our approach can identify the most essential feedback states for locomotion skills, including balance recovery, trotting, bounding, pacing and galloping. By using only key states -- joint positions, gravity vector, base linear and angular velocities -- we demonstrate that a simulated quadruped robot can achieve robust performance in various test scenarios across these distinct skills. The benchmarks using task performance metrics show that locomotion skills learned with key states can achieve comparable performance to those with all states, and the task performance or learning success rate will drop significantly if key states are missing. This work provides quantitative insights into the relationship between state observations and specific types of motor skills, serving as a guideline for robot motor learning. The proposed method is applicable to differentiable state-action mapping, such as neural network based control policies, enabling the learning of a wide range of motor skills with minimal sensing dependencies.

\pdfbookmark[1]{Introduction}{Introduction}
\section*{Introduction}
The notion of learning machines predates the origins of cybernetics, control theories and apparatus in 1940s \cite{wiener2019cybernetics}, with a longstanding interest in creating functioning replicas of living organisms. Robots with morphologies similar to their biological counterparts provide unique opportunities to develop machines with motion capabilities comparable to that of animals. As easy-to-control platforms, robots allow scientists to study sensorimotor learning, providing opportunities to conduct control experiments and generate quantitative data analysis \cite{ijspeert2014biorobotics,karakasiliotis2016cineradiography, nyakatura2019reverse,cheng2020neuroengineering}.

In the field of robotics, a large part of physical motor skills can be formulated as feedback control, i.e., control policies represented as state-action mapping. For conventional model-based approaches, such as trajectory optimization, task skills are represented by motion trajectories which are optimized based on the prior models from the domain knowledge \cite{neunert2018whole,kim2019highly}, and the resulted motions are the natural outcomes of the optimized physical interactions rather than being the underlying motor control policies. For learning-based approaches, motor policies in the form of Deep Neural Networks (DNNs) can perform closed-loop control and generate actions constantly based on the sensing of feedback states. The selection of feedback states is essential for learning robot skills effectively, because inappropriate feedback states can impede the learning process, especially if key feedback is missing, then the robot will not be able to achieve the desired behaviors or will perform very poorly. Physics simulation allows access to as many ideal feedback states as possible, potentially leading to better results \cite{kalashnikov2018qt}. However, while transferring learned policies to real systems, often the easily-available privileged information in the simulation are no longer accessible or not reliable in real-world applications. Also, some states are not directly measurable in real robots compared to that in the simulation, and therefore need to be estimated by state estimation which is subject to computational errors and uncertainties \cite{xie2020dynamics,ibarz2021train}. In conclusion, it is desirable to employ realistic sensory information and avoid using unreliable or inaccessible sources of feedback during the training of DRL-based policies. 

Deep reinforcement learning (DRL) is one of the promising approaches to learn a wide range of whole-body motor skills in robotic applications. For example, DRL methods can learn different legged locomotion skills, such as trotting \cite{cite:hwangbo2019learningAgile,lee2020learning,peng2020learning,yang2020multi}, pacing, spinning \cite{peng2020learning}, walking \cite{haarnoja2018learning}, galloping \cite{tan2018sim}, balance recovery \cite{cite:hwangbo2019learningAgile}, and multi-skill locomotion \cite{yang2020multi}. Although deep neural networks (DNNs) show success in acquiring complex motor skills, due to the lack of interpretability \cite{yosinski2015understanding,jimenez2020drug}, it is yet to answer how to determine the relative importance of different feedback states, or what state observations are more effective than others. Therefore, it often requires a lot of human insights and engineering efforts to select appropriate feedback states empirically in robot learning \cite{tassa2018deepmind,kalashnikov2018qt,reda2020learning,lee2020learning,marasco2021neurorobotic, thandiackal2021emergence,lee2020making}.

Different recent work selects a combination of representative feedback signals for learning quadruped locomotion. For example, joint position, joint velocity, angular velocity, and body orientation are used to learn free gait transitions between walking, trotting, pacing and bounding \cite{shao2021learning}. History of feedback signals has also been used in existing work, where a quadruped robot can adjust locomotion policies in the real world using history of root orientation, joint position and previous actions as input \cite{smith2022legged}. High speed locomotion on natural terrains has been achieved by using joint position and velocity, body orientation (gravity vector), and previous actions \cite{margolis2022rapid}. Overall, these feedback states are selected empirically and tested by trials without a systematic approach to determine the qualitative or quantitative importance of various feedback states among all the input signals.

\pdfbookmark[2]{Biological insights and motivation}{Biological insights and motivation}
\subsection*{Biological insights and motivation}
\begin{figure}[t]
\centering
\includegraphics[width=0.9\textwidth]{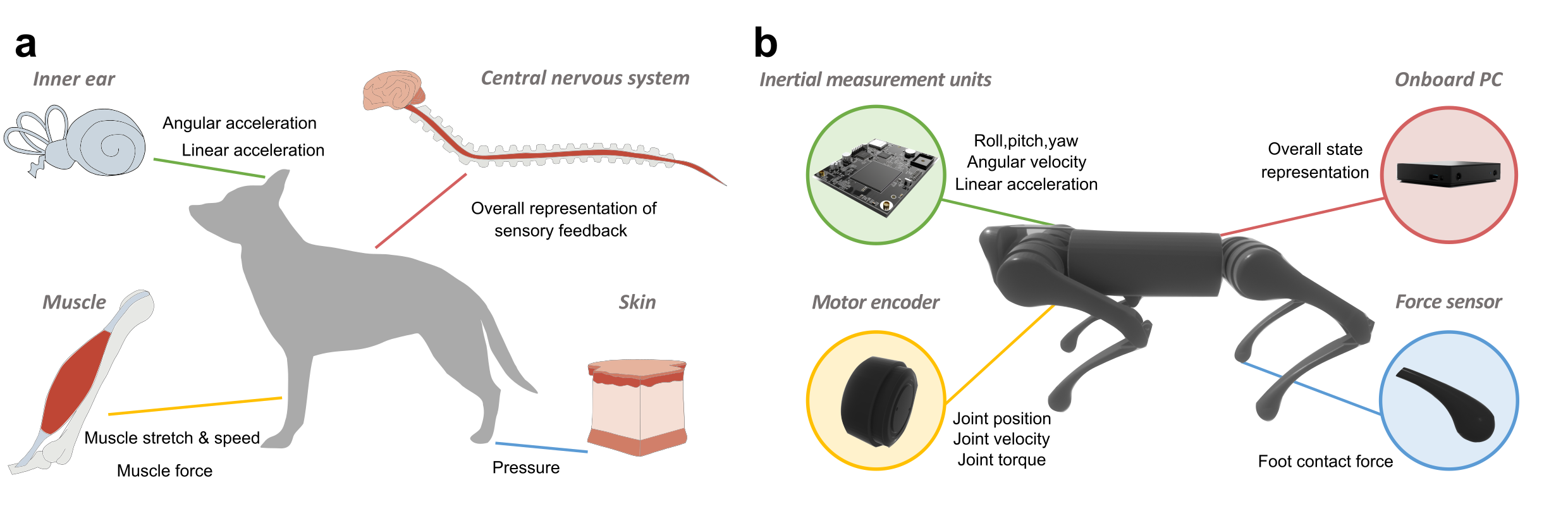}
\caption{\textbf{Comparison of various functional sensory feedback signals between the quadrupedal animals and their robotic counterparts.} \figcap{A} Sensory organs and feedback of a dog. The central nervous system fuses sensory information from various organs, such as the inner ear, muscle and skin, and then produces motor commands. The vestibular system of vertebrates senses angular and linear accelerations, muscle spindles measure the stretch and stretch speed of skeletal muscles, the Golgi tendon organs measure exerted muscular forces, and the skin feels pressures among others. \figcap{B} Sensors and feedback on a legged robot. An onboard computer processes measurements of signals from different sensors, such as an inertial measurement unit, motor encoders and force sensors, and then generates actions for electric motors.}
\label{fig:biology}
\end{figure}

Biological studies find that animals use multimodal sensory information collected from various sensory organs to achieve different locomotion tasks, including visual, mechanical, chemical, and thermal sensation, all in unison to render feedback from their own movements and the surroundings \cite{dickinson2000animals,rossignol2006dynamic,taylor2007sensory,carpenter2012neurophysiology}. As shown in Fig. \ref{fig:biology}\capA, the vestibular system of vertebrates senses angular and linear accelerations, muscle spindles measure the stretch and stretch speed of skeletal muscles, the Golgi tendon organs measure exerted muscular forces, and the skin feels pressures among others. All this sensory information inherits redundancy which ensures robustness for the movement control in animal locomotion \cite{roth2016integration}. The Central Nervous System (CNS) fuses all information from these sensory organs and generates appropriate motor actions to complete tasks. To study the importance of different sensing information from various sensory organs, biological studies use lesion studies or ablation studies \cite{cox2018influence,sober2005flexible}. However, it is challenging to conduct such experiments on live animals due to ethical limitations and the difficulty of selectively stimulating and/or removing different receptors \cite{pearson2006assessing}. To date, robotic systems can have onboard sensors similar to their biological counterparts (see Fig. \ref{fig:biology}), and therefore offer a range of comparable proprioceptive and exteroceptive information to learn animal-like motor skills. The biological findings indicate that to recreate similar sensorimotor control in mechatronic robots, the understanding of the importance of different sensory feedback is crucial in order to produce desired behaviors, which is still missing in the robotics field. 

\pdfbookmark[2]{Related work}{Related work}
\subsection*{Related work}
Ablation experiments are commonly adopted to study the qualitative importance of individual feedback states, which focus on the impact of removing a single feedback state on performance. Experiments on a lamprey-like robot demonstrated that the distributed hydrodynamic force feedback contributes to the generation and coordination of rhythmic undulatory swimming motion \cite{thandiackal2021emergence}. Ablation studies showed that including the Cartesian joint position or contact information has different influences on learning robot locomotion behaviors \cite{reda2020learning}. Three sets of proprioceptive state feedback are used to learn CPG-based quadruped locomotion separately and foot contact booleans and CPG states are finally selected through ablation studies \cite{bellegarda2022cpg}. In the above research using ablation studies, a feedback state is considered to be important if the performance drops after it has been removed. In other words, the ablation study examines the difference of performance with and without the signals of interest. Therefore, the above ablation methods are not able to provide a quantitative ranking of the relative importance of each feedback signal, or the percentage of contribution of one sensing signal in comparison with others among the whole set of sensory feedback.

There have also been limited attempts to compare the importance of a certain type of feedback at different time steps quantitatively. For example, the influence of proprioceptive states on foot height commands was quantified and compared at different time steps \cite{lee2020learning}. However, this quantitative approach is used for analyzing the foot-trapping behavior during trotting and has not been extended to studying other motor skills.

To the best of our knowledge, the literature suggests a gap in formalizing systematic and quantitative approaches to compare the relative importance of different sensory feedback, which are the enabling tools to select the essential feedback and to learn various robot skills more effectively. Therefore, our research aims to answer the following open questions in the context of robot learning: What is the relative importance of different sensory feedback signals for a given motor task and various quadrupedal gaits? Which sensory feedback signals are essential, necessary and sufficient for learning quadrupedal locomotion? Which redundant feedback is beneficial to have but not entirely necessary?

\pdfbookmark[2]{Contribution}{Contribution}
\subsection*{Contribution}
Here we propose a systematic approach to study the quantitative influence of different sensory feedback among distinct quadrupedal tasks, i.e., balance recovery, trotting and bounding, which are learned as feedback control policies represented by neural networks. We rank the level of importance of sensory feedback and identified a common set of essential feedback states for general quadruped locomotion. Further, we apply the proposed approach to learn new locomotion skills such as pacing and galloping using only the most essential feedback states.

In summary, our main contributions are: (i) Development of a systematic saliency analysis method to specifically quantify and rank the importance of each sensory feedback for a specific motor task; (ii) Identification of a common set of essential feedback states for general quadruped locomotion based on the saliency analysis of representative locomotion skills; (iii) Successful robot learning of new motor skills using only essential feedback states, demonstrating the efficacy of a minimal set of sensors. 

Our study contributes to identifying the most essential feedback, i.e., key states, in a task-specific manner, enabling robust motor learning using only the key states. The results provide new insights into the quantitative relative importance of different feedback states in locomotion behaviors. The identification of essential sensory feedback guides the selection of a minimum and necessary set of sensors, allowing robots to learn and perform robust motor skills with minimal sensing dependencies.

\pdfbookmark[1]{Results}{Results}
\section*{Results}
This section elaborates on the investigation of the quantitative relative importance of common feedback states for three representative and distinct locomotion skills (balance recovery, trotting and bounding), and identification of a set of key feedback states that are consistently more important than others: joint positions, gravity vector (i.e., body orientation), base linear and angular velocities. Our results show that learning locomotion skills with only the key feedback states achieves comparable performance to using all available states. Further, we demonstrate the effectiveness of these key feedback states for learning new locomotion skills, such as pacing and galloping. Finally, we validate the robustness of the key-state policies in scenarios that were not encountered during training, and against robot uncertainties including sensing noises, and variations of robot mass and control gains. The concepts and ideas of our approaches are presented in Fig. \ref{fig:flowchart} and more results can be seen in Supplementary Video 1-4.

\begin{figure}[H]
\centering
\includegraphics[width=0.9\textwidth]{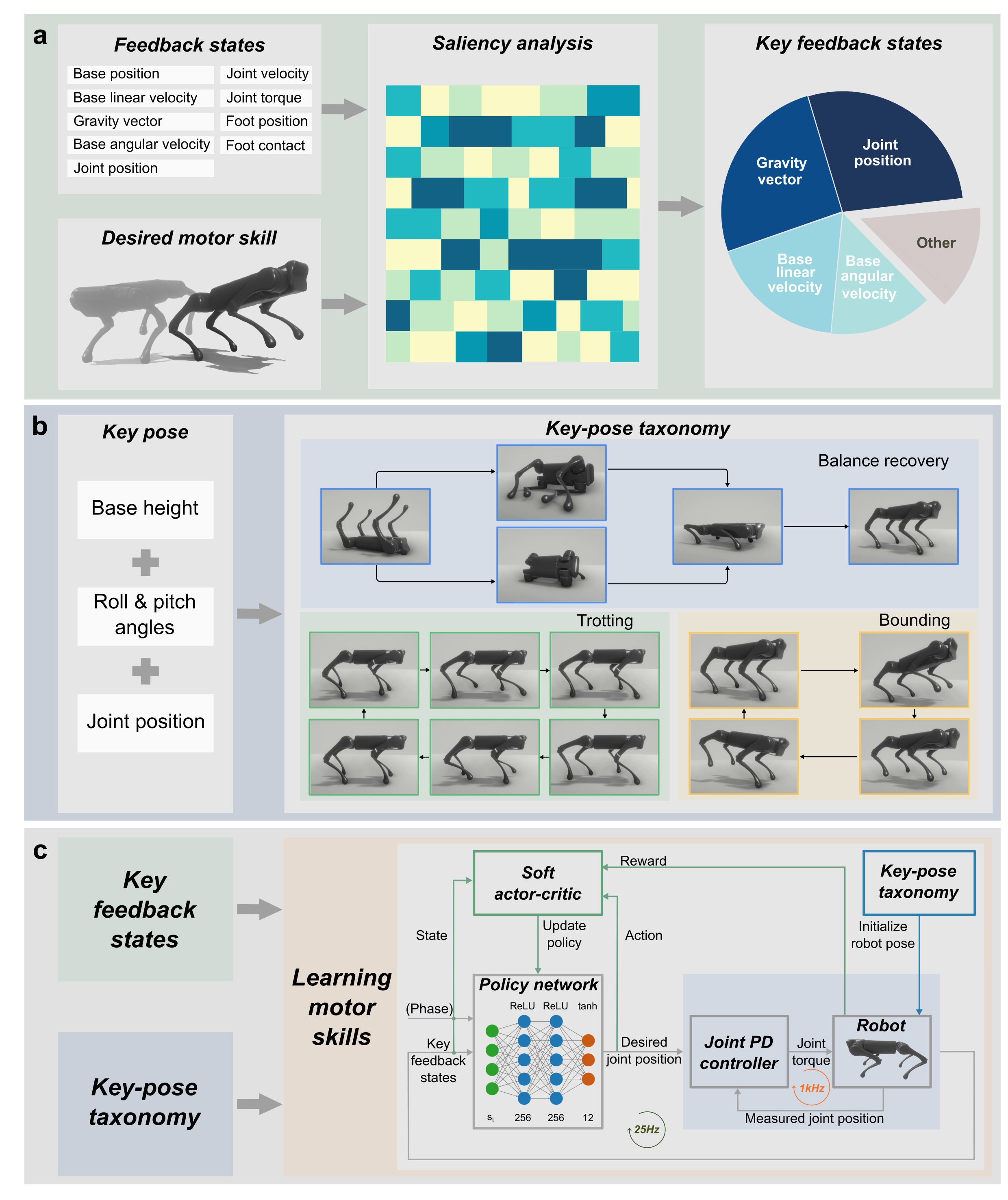}
\caption{\textbf{Proposed approach for identifying and employing key feedback states to learn effective locomotion skills.} \figcap{A} The proposed saliency analysis can rank the importance among different feedback states, given a set of feedback states and a targeted task-level skill. The doughnut chart shows our findings on the essential feedback states for quadruped locomotion, including joint position, gravity vector, base linear velocity and base angular velocity. \figcap{B} Key-pose taxonomy used in robot pose initialization for learning effective balance recovery, trotting and bounding, respectively. Each key pose is represented by an array of body height, body orientation (roll and pitch angles) and joint positions, which determines the pose of a floating-base robot and is categorized by distinct contact patterns that are unique to the targeted type of gait. \figcap{C} The DRL framework utilizing key feedback states and key-pose taxonomy that are sufficient for successful and effective learning of robot motor skills, where the phase vector $(sin2\pi\phi, cos2\pi\phi)$ inputs to the policy network in parallel with key feedback states for periodic locomotion skills ($\phi$ is temporal information representing 0-100\% phase over a gait period).}
\label{fig:flowchart}
\end{figure}

\pdfbookmark[2]{Identifying key feedback states for quadruped locomotion}{Identifying key feedback states for quadruped locomotion}
\subsection*{Identifying key feedback states for quadruped locomotion}

\pdfbookmark[3]{Quantifying the relative importance of feedback states}{Quantifying the relative importance of feedback states}
\subsubsection*{Quantifying the relative importance of feedback states}
We formulate a systematic saliency analysis for quantifying the relative importance of various feedback states for a desired motor skill. Inspired by the commonly available sensory information in animals (see Fig. \ref{fig:biology}\capA) and those widely used in deep reinforcement learning of robot locomotion \cite{yang2020multi, lee2020learning}, we design a common set of feedback states, referred as full feedback states, including nine types of feedback states with 64 dimensions in total: base position, gravity vector, base angular velocity, base linear velocity, joint position/angle, joint velocity, joint torque, foot position, and foot contact (contact status or forces) (see Methods for definitions). Using this full set of states, we obtain the neural network policies for locomotion skills on the \textit{A1} quadruped robot \cite{unitree} in PyBullet simulation via the DRL framework detailed in Fig. \ref{fig:flowchart}\capC~ and Supplementary Note 1. At each time step, we compute the saliency values of each dimension of the feedback states with respect to the associated actions using the integrated gradients method \cite{sundararajan2017axiomatic} (see Methods). The saliency value measures the influence of the input signal on generated actions. For each feedback dimension, it quantifies the amount of changes in output actions as the input signal varies at a certain time step. The higher the saliency value, the more associated actions change as the input signal varies, indicating a higher level of importance and task-relevance of a particular signal. Finally, we formulate the \textit{relative importance} of a given feedback state as the percentage of its quantified importance over the entire period of motions with respect to the sum of importance of all the feedback states (see Methods).

\begin{figure}[H]
\centering
\includegraphics[width=0.9\textwidth]{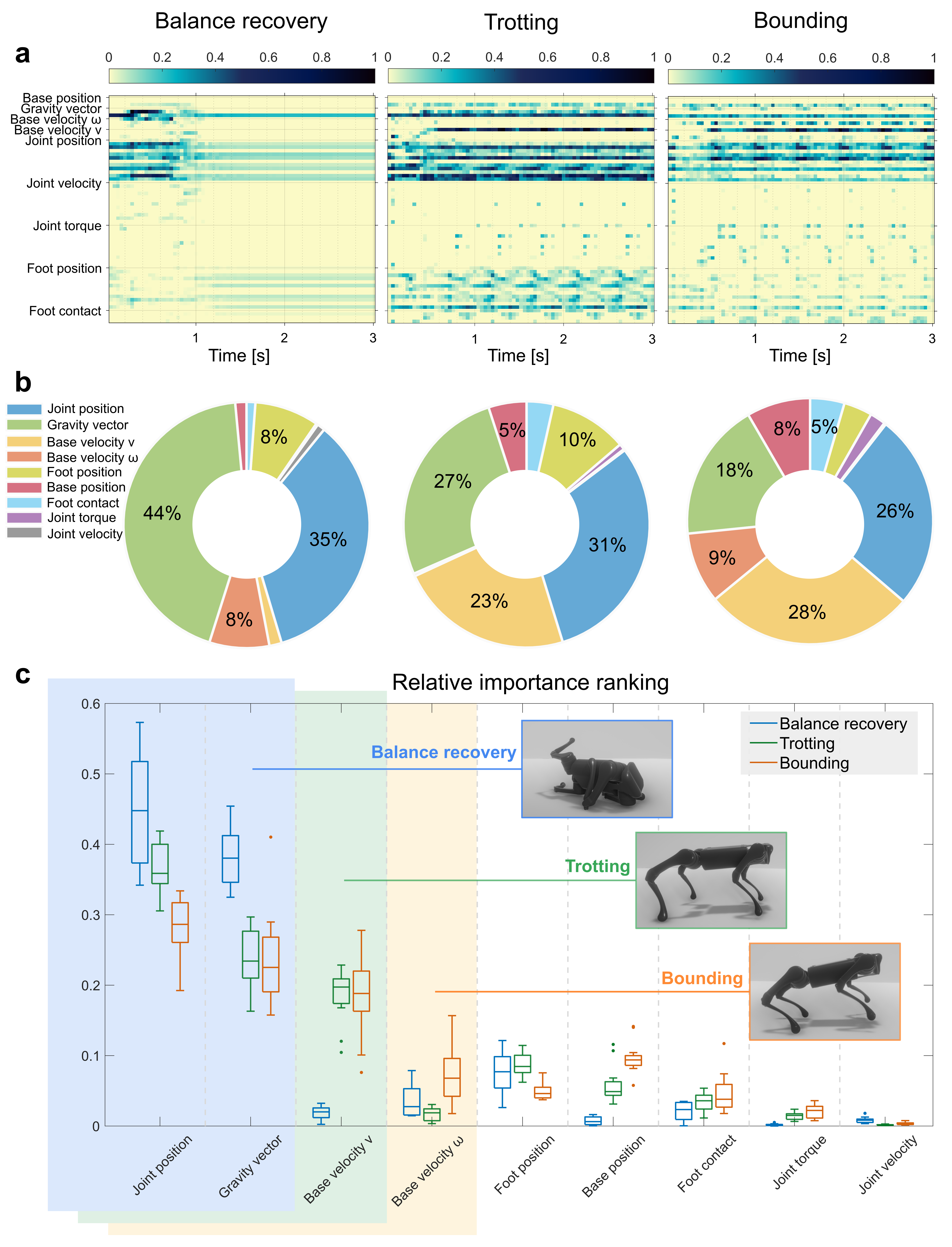}
\caption{\textbf{Ranking of feedback states and saliency analysis that identified top four key states for balance recovery, trotting and bounding skills.} \figcap{A} Saliency maps showing the importance of the 64-dimensional feedback for three learned skills, respectively. At each time instant in the saliency map, a darker pixel indicates that the corresponding feedback signal has more importance and influence on the generated actions, compared to others. \figcap{B} Doughnut charts showing the relative importance of nine feedback states for three learned skills, summarizing the overall averaged importance based on Fig. \ref{fig:saliency}\capA. \figcap{C} A box plot showing statistics of relative importance of each feedback state for balance recovery, trotting, and bounding, respectively. Each box reflects the slight variation of the relative importance of the corresponding feedback state among multiple learned policies for each locomotion skill (see details in Supplementary Figure \ref{saliency-stand}-\ref{saliency-boxplot}). The full set of nine feedback states are ranked in an order of relative importance from high to low, suggesting the key feedback states for quadruped locomotion include joint positions, gravity vector (i.e., body orientation), base linear velocity, and base angular velocity. Blue, green and yellow shaded areas enclose key feedback states for three locomotion skills, respectively.}
\label{fig:saliency}
\end{figure}

\pdfbookmark[3]{Key feedback states for quadruped locomotion}{Key feedback states for quadruped locomotion}
\subsubsection*{Key feedback states for quadruped locomotion}\label{sec:key-state}
Here we identify key feedback states from the collection of nine feedback states according to the ranking of relative importance for three representative and distinct locomotion tasks: balance recovery, trotting and bounding, respectively. 

To assess the significance of each feedback state in relation to their associated actions, we visualize saliency values as saliency maps (see Fig. \ref{fig:saliency}\capA) for highlighting their relative importance, as well as the summary using doughnut charts (Fig. \ref{fig:saliency}\capB) to delineate the contribution of each state over time and their overall impacts to the above skills -- which reveals key feedback states with around 80\% relative importance in total for the three skills summarized in Fig. \ref{fig:general}\capC, compared to 20\% relative importance for task-irrelevant states. 

From the time plots of relative importance for nine feedback states in Extended Data Figure \ref{ex:saliency-time}, we found that relative importance varies depending on the robot posture or phase over time. During balance recovery, gravity vector and joint positions are found to be the most important feedback state during body flipping and standing, respectively. During periodic trotting and bounding, within each gait cycle, the most crucial feedback state alternates between joint positions and base linear velocity.

\begin{figure}[H]
\centering
\includegraphics[width=0.9\textwidth]{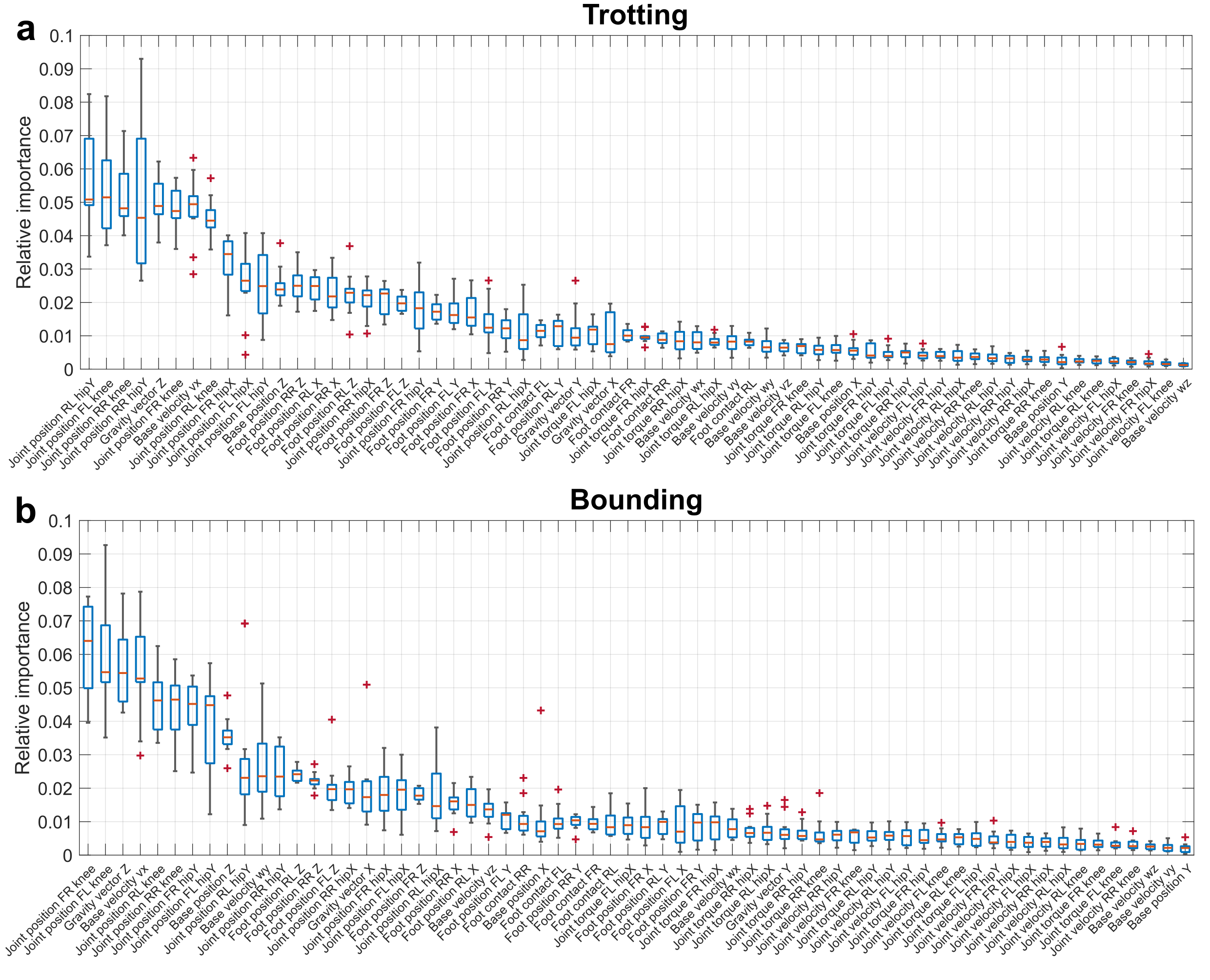}
\caption[Comparison of the relative importance over 64 dimensions of feedback signals for trotting and bounding.]{Comparison of the relative importance of 64 dimensions of feedback state for trotting and bounding. Each box shows the median (red horizontal line), 25th and 75th percentiles (lower and upper blue horizontal lines), minimum and maximum (lower and upper gray horizontal lines), and outliers (red plus sign) of the relative importance of the corresponding state dimension from twelve different trials for the corresponding locomotion task. \figcap{A} A boxplot expanding the boxplot in Supplementary Figure \ref{saliency-boxplot}b, which shows relative importance ranking among 64 state dimensions for trotting on a flat ground. \figcap{B} A boxplot expanding the boxplot in Supplementary Figure \ref{saliency-boxplot}c, which shows relative importance ranking among 64 state dimensions for bounding on a flat ground.}
\label{fig:saliency-dim}
\end{figure}

Our study reveals that each joint has a different level of contribution to distinct locomotion tasks (Fig. \ref{fig:saliency-dim}). For example, sagittal movement states and joints that have a large range of motions usually have a higher contribution to locomotion than others. During trotting, as the rear legs deliver more power to propel the robot forward and overcome energy loss caused by friction and impacts, the rear hip pitch joints move in larger ranges, resulting in higher importance. During bounding, the front knee joints are more important than the rear knee joint positions, as bounding requires the front knee joints to buffer landing impacts more and provide stable body weight support during the pre-landing and stance phase.

Overall, through our proposed systematic approach, a unified set of key states can be identified across three representative and distinct locomotion skills for quadruped robots, which are joint position, gravity vector, base linear velocity and base angular velocity.

\pdfbookmark[3]{Key feedback states under various circumstances}{Key feedback states under various circumstances}
\subsubsection*{Key feedback states under various circumstances}\label{sec:cases}

\pdfbookmark[4]{Rough terrain}{Rough terrain}
\paragraph{Rough terrain}

In previous results, we train the trotting and bounding on a flat ground. Here we aim to investigate if training on rough terrain will cause any changes in key feedback states. We train new trotting and bounding policies on rough terrain in a $6.4 m \times 6.4m$ area, which consists of $4096$ cubes each with 0.1 m length and width, and heights sampled from $(0\sim3cm)$. Results in Extended Data Figure \ref{ex:boxplot-rough} show that the key states for trotting and bounding remain the same. 

\pdfbookmark[4]{Gait frequencies}{Gait frequencies}
\paragraph{Gait frequencies}

Trotting and bounding policies were trained with 1 Hz, 2Hz and 5 Hz, respectively. Saliency maps and doughnut charts in Supplementary Figure \ref{frequency} indicate that key feedback states for trotting and bounding are consistent across low, medium and high gait frequencies. Thus, we conclude that the identified key feedback states are not affected by the gait frequency within such a range.

\pdfbookmark[4]{Foot contact status versus foot contact forces}{Foot contact status versus foot contact forces}
\paragraph{Foot contact status versus foot contact forces}

The foot contact can be represented in various formulations for learning locomotion skills. The previous analysis uses sigmoid contact which is a continuous signal indicating the contact status (see Methods). Here we replace the sigmoid foot contact with normalized foot contact forces for learning trotting and bounding, where the foot contact forces are normalized by the body weight and capped between zero and one. We conclude that using either of the foot contact formulations as feedback states renders the same conclusions regarding the importance of foot contact (see Supplementary Figure \ref{contact}).

\pdfbookmark[3]{Importance of history states}{Importance of history states}
\subsubsection*{Importance of history states}
To investigate the impact of historical information on sensory feedback selection, we trained a new policy for balance recovery with an expanded set of state input, including states at the current time step and two history steps. The results of the saliency map and the bar plot in Extended Data Figure \ref{ex:history}\capA-\capB ~show that: (i) for each type of feedback state, current information is more important than history information; (ii) important states remain important within each set of history states; and (iii) the overall importance of all states at the current time step is much higher than that at both history steps.

\pdfbookmark[3]{Correlation between feedback states}{Correlation between feedback states}
\subsubsection*{Correlation between feedback states}
Heatmaps were generated to reflect the absolute value of Pearson correlation coefficient \cite{cohen2009pearson} across feedback states (Fig. \ref{fig:general}\capE~ and Extended Data Figure \ref{ex:history}\capC). The average correlation coefficients between any two types of current feedback states were visualized using chord diagrams in Fig. \ref{fig:general}\capF ~and Extended Data Figure \ref{ex:history}\capD. These findings reveal that the key states identified for balance recovery are correlated with all other task-irrelevant states to varying degrees. Moreover, the measurements of these key states are usually less noisy, making them more suitable to be used. Therefore, these results suggest that selecting key states with underlying correlations with other signals can effectively reduce the number of sensors required for feedback in the closed-loop control.

\pdfbookmark[3]{Quantifying the relative importance of the feedforward phase vector}{Quantifying the relative importance of the feedforward phase vector}

\subsubsection*{Quantifying the relative importance of the feedforward phase vector}\label{sec:phase}
\pdfbookmark[4]{Importance of the feedforward phase vector}{Importance of the feedforward phase vector}
\paragraph{Importance of the feedforward phase vector}

For trotting and bounding, the phase vector is used to enforce the cyclic pattern as a \textit{feedforward} input to the neural network besides the feedback states (Fig. \ref{fig:flowchart}\capC~ and Methods). We shall note that the phase vector is included in the training of all feedback control policies for trotting and bounding but excluded from the ranking of the important states except this section. Without the phase vector as input, the robot would fail to learn cyclic motion. The analysis in Extended Data Figure \ref{ex:phase1} shows that the feedforward phase vector counts for the relative importance of 28\% and 38\% for trotting and bounding, respectively, which is more important than any type of feedback states.  

\pdfbookmark[4]{Importance of the phase vector during swing and stance}{Importance of the phase vector during swing and stance}
\paragraph{Importance of the phase vector during swing and stance}

From the time plot of saliency value for phase vector in Extended Data Figure \ref{ex:phase2}, we found the importance of the phase vector follows a cyclic pattern, reaching the highest value twice within each gait around the transitions between swing and stance during trotting and bounding. The cyclic pattern of the phase vector is synchronized with foot-ground contact. Such high importance during the contact transitions indicate that the phase vector regulates the timing of establishing and breaking foot-ground contact, resulting in a synchronized cyclic pattern with the transitions of swing and stance.

\pdfbookmark[2]{Benchmarking of learned motor skills}{Benchmarking of learned motor skills}
\subsection*{Benchmarking of learned motor skills}\label{sec:benchmark}
To validate the effectiveness of our proposed approach to identifying important states, we formulate task-related  performance metrics for the quantitative evaluation of three locomotion tasks (see Methods), and benchmark the following five settings in ten scenarios with the same DRL framework: (i) \textit{full-state policies} with nine feedback states (see Fig. \ref{fig:saliency}\capA) and random robot pose initialization for training; (ii) \textit{key-state policies} with four key states (see Fig. \ref{fig:general}\capC) and key-pose initialization; (iii) irrelevant-state policies which only use five less important states; (iv) open-loop trajectories from full-state policies; and (v) open-loop trajectories from key-state policies. Open-loop trajectories repeat the desired joint positions for two gait periods generated by the feedback policies. 

\begin{figure}[H]    
    \centering  
        {\includegraphics[width=0.9\textwidth]{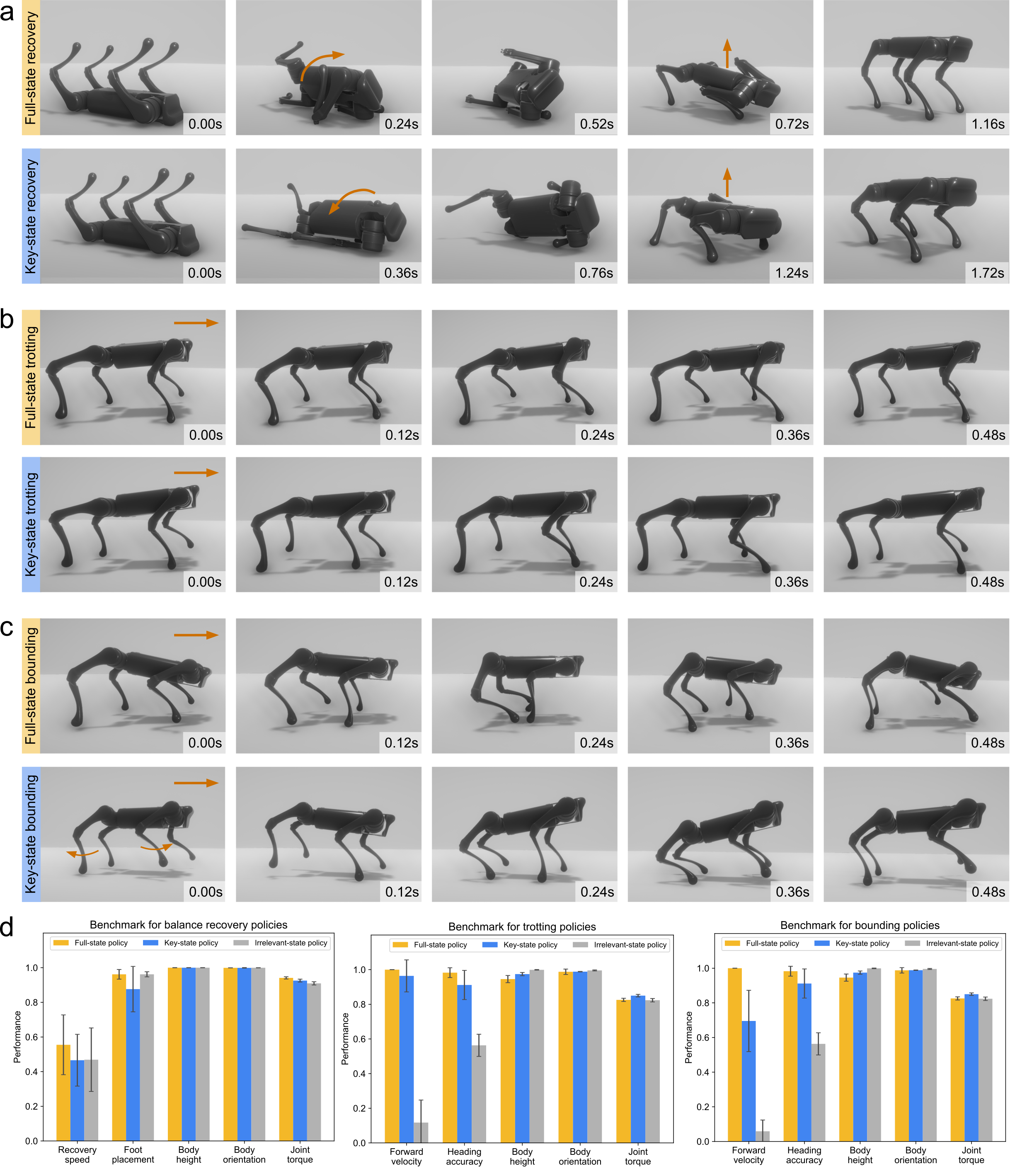}}
    \caption{\textbf{Comparison of task performance for full-state policies and key-state policies for balance recovery, trotting and bounding.} \figcap{A} Balance recovery from lying down on the back by a full-state policy (top) and a key-state policy (bottom). \figcap{B} Learned trotting gait by a full-state policy (top) and a key-state policy (bottom). \figcap{C} Learned bounding gait by a full-state policy (top) and a key-state policy (bottom). \figcap{D} Comparison of task performance using metrics for full-state, key-state, and irrelevant-state policies over balance recovery, trotting, and bounding. Higher values of metrics indicate better performance. The key-state policies achieved 94.7\%, 99.1\% and 93.7\% on average, respectively, compared to that of the full-state policies. Irrelevant-state trotting and bounding policies achieve 11.7\%, 5.9\% in forward velocity and 57.3\%, 57.3\% in heading accuracy with respect to full-state policies. Although the selected irrelevant-state balance recovery policy achieves similar performance to full-state and key-state policies, the success rate of learning such policy drops significantly. Details of robustness tests in Supplementary Note 3.}
    \label{fig:benchmark}    
\end{figure}

For balance recovery, the metrics include recovery speed, final foot placement, final body height, final body orientation and joint torque. For trotting and bounding, the same physical quantities are used, including forward velocity, heading accuracy, body height, body orientation and joint torque. By averaging the five performance metrics (see Methods), we found the performance of key-state policies is comparable to full-state policies for three skills (Fig. \ref{fig:benchmark}\capD), and if key feedback states are missing, there would be a significant drop in task performance (forward velocity and heading accuracy for trotting and bounding) or learning success rate (balance recovery). More details can be found in Fig. \ref{fig:benchmark}\capA-\capC, Supplementary Figure \ref{benchmark-detail} and Supplementary Note 2. 

Furthermore, the performance benchmark of the closed-loop control policies versus the open-loop trajectories (Supplementary Figure \ref{benchmark-std}-\ref{benchmark-std-3cm}) demonstrates the importance of utilizing feedback states to correct robot behaviors. Also, compared to the open-loop trajectories from the full-state policies with random explorations, results indicate that trajectories learned by the key-state policies are typically more stable when executed in an open-loop manner. This is because the neural network tends to discover more conservative and stable trajectory patterns surrounding the solution space initialized by the key poses. 

\pdfbookmark[2]{Applicability to learning new skills}{Applicability to learning new skills}
\subsection*{Applicability to learning new skills}
\begin{figure}[H]
\includegraphics[width=0.9\textwidth]{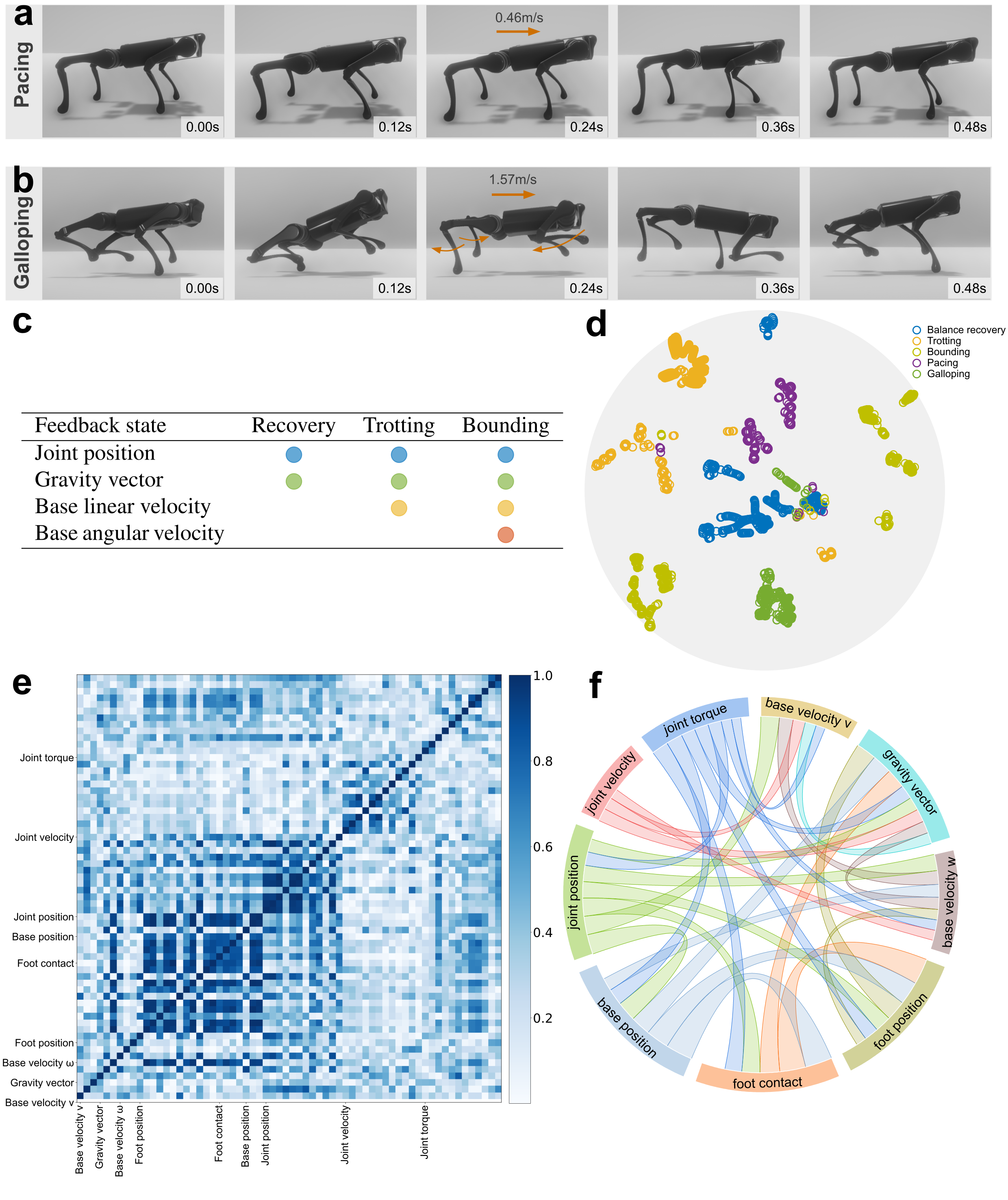}
\caption{\textbf{The use of key feedback states for learning novel locomotion skills.} \figcap{A} Successful pacing gait learned by \textit{A1} robot with an average forward velocity of $0.46m/s$. \figcap{B} Successful galloping gait learned by \textit{A1} robot with an average forward velocity of $1.57m/s$. \figcap{C} Key feedback states used for balance recovery, trotting and bounding skills. Each color represents one type of feedback state: blue, green, yellow and orange dots represent joint position, gravity vector, base linear and angular velocities, respectively. \figcap{D} The 2D projections of the 18-dimensional trajectories (body height, body linear velocity, roll and pitch angles and joint positions) sampled from key-state pacing and galloping policies, full-state and key-state balance recovery, trotting and bounding policies using t-SNE. \figcap{E} Heatmap showing absolute Pearson correlation coefficient between any two dimensions of nine feedback states from full-state balance recovery policy, where a darker color indicates stronger correlation. \figcap{F} Chord diagram summarizing the correlation between any two feedback states for full-state balance recovery (self-correlation and percentage $<25\%$ are removed for clarity), where the wider the link between the two states, the stronger they correlate with each other.}
\label{fig:general}
\end{figure}

This section aims to investigate whether the identified key feedback states can be directly used for learning new locomotion skills. Using the key feedback states identified from three representative locomotion skills and the newly designed task-specific key poses (Supplementary Figure \ref{taxonomy}), the \textit{A1} quadruped robot successfully learns robust pacing and galloping gaits, (Fig. \ref{fig:general}\capA-\capB~ and Supplementary Video 4). 

Based on the t-Distributed Stochastic Neighbor Embedding (t-SNE) \cite{van2008visualizing} plot in Fig. \ref{fig:general}\capD, we found that the trajectories of key-state pacing and galloping are located within the circle formed by the trajectories of balance recovery, trotting and bounding policies. This suggests that the studied skills encapsulate the common patterns of leg movements with sufficient diversity and the newly learned skills are closely-related neighborhood skills to these skills. Therefore, using the same set of key states can effectively learn both new gait patterns. However, learning more distinct new locomotion skills may require identifying new key feedback states, which can be achieved by reapplying our approach.

\pdfbookmark[1]{Discussion}{Discussion}
\section*{Discussion}
This work has developed a quantitative analysis method to investigate the quantitative influence of various common feedback states for learning locomotion skills. In contrast to qualitative ablation experiments for state selection, the quantitative importance ranking obtained by our proposed systematic and efficient approach offers a useful guideline for selecting the most essential and task-relevant feedback states to learn a wide range of robot behaviors, such as balance recovery, trotting, bounding, pacing, and galloping. As a result of motor learning through neural networks, the importance of states is indirectly encapsulated by a large amount of learned neural network weights. 

Our method contributes to the interpretation, comparison and validation of the state importance by a direct ranking of quantitative relative importance over common sensory feedback for quadruped locomotion. The study suggests that joint positions, gravity vector, base linear and angular velocities are the essential states, comprising 80\% total relative importance for motor learning. In comparison, foot positions, foot contact, base position, joint torque and velocities play a less critical role and are the types of feedback signals that are better to have but not entirely indispensable, and therefore they can be excluded from motor learning without significantly affecting robot performance.

\pdfbookmark[2]{Benefits for robot control design}{Benefits for robot control design}
\subsection*{Benefits for robot control design}
Our method provides a quantitative ranking of feedback states and hence identifies the level of their importance, guiding the selection of a minimum and suitable set of sensors to learn robust quadruped locomotion. Different combinations of sensors can be chosen based on the hardware availability of particular applications. For example, we found motor encoders, Inertial Measurement Unit (IMU), and estimation of body linear velocities suffice to achieve common quadruped locomotion skills. 

To summarize, this research benefits the design of robot control in several aspects:
(i) improves design efficiency of the learning framework by selecting important feedback states through one single training session, which is more efficient than empirical trial-and-error or ablation studies that require multiple iterative processes;  
(ii) promotes lightweight and cost-effective robot design by equipping only task-relevant sensors;
(iii) reduces the need for developing state estimation for task-irrelevant or unimportant states, thus reducing the dependencies of task success on sensing and state estimation uncertainties that helps to mitigate sim-to-real mismatch.

Our quantitative analysis approach requires a successful sensorimotor policy, i.e., a differentiable state-action mapping of the motor skills, to be used for identifying the importance level of states. In cases where motion learning is initially infeasible via reinforcement learning, all commonly available sensing shall be included to facilitate motor learning, or other approaches like supervised learning or imitation learning can be used if demonstrations are available.

\pdfbookmark[2]{Feedforward pathways}{Feedforward pathways}
\subsection*{Feedforward pathways}
In this work, to simplify the implementation, the periodic feedforward signals are implemented as a 2-dimensional phase vector $(sin2\pi\phi, cos2\pi\phi)$ generated directly using the temporal information $\phi$ (0-100\% phase over a gait period), representing continuous periodicity (see Fig. \ref{fig:flowchart}\capC). The use of such phase input serves as a representation of priors or prior knowledge as in other previous works \cite{jonschkowski2014state, yang2020learning}, and the phase signals are not modulated by sensory feedback signals in our study. 

The use of the feedforward signals, i.e., the 2-dimensional phase vector, may result in a small contribution of foot contact information as revealed in this study. The types of locomotion skills investigated in this study do not involve complex terrain properties (e.g., soft, muddy, granular), and the limited uncertainties of the ground contact further reduce the necessity or reliance on foot contact forces during the motor learning process of state-action mapping. Additionally, our supplementary analysis has revealed a correlation between foot contact and the phase signals, as detailed in Supplementary Figure \ref{phase-foot}. This correlation also implies that once the locomotion gait reaches a limit cycle, a stable gait can be sustained without the explicit use of foot-ground contact information if there are no large external perturbations. 

In principle, these feedforward signals could be implemented in CPG networks and be modulated in various ways, e.g. with phase resetting mechanisms or with feedback terms that continuously modulate the phase and amplitude of CPG signals, as in \cite{owaki2013simple,aoi2006stability,fujiki2018adaptive,bellegarda2022cpg}. In future work, it would be interesting to include such feedback mechanisms and investigate whether the relative importance of different sensory modalities would change. For instance, it might lead to a higher importance of load feedback, which has been suggested to be important for cat locomotion \cite{ekeberg2005computer} and shown to be a sufficient source of information for interlimb coordination \cite{owaki2013simple}.

\pdfbookmark[2]{Relation to biology}{Relation to biology}
\subsection*{Relation to biology}
On top of our findings contributing to robotics and learning, the importance of key feedback states provides meaningful implications on sensorimotor control in animal locomotion. In general, our findings are in agreement with biological findings and hypotheses. The key sensory feedback found by our studies on quadrupedal robots maps to the vestibular system and muscle spindles which have been proved to be critical for postural control and goal-directed vertebrate locomotion on the biological counterparts \cite{rossignol2006dynamic,grillner2008neural}. Our analysis also reveals that important states vary with tasks and certain sensory signals are more critical than others at different moments during a gait cycle, consistent with existing biological findings and hypotheses \cite{rossignol2006dynamic,caggiano2018midbrain}. 

It is important to note that our conclusions on important states stem from common locomotion skills on a mechanical robot and may differ from the general findings in animals, such as the importance of contact forces and limb loading as discussed in the previous section. Our contribution to biology is to provide biologists with the computational approach to identify important states on simulated animals, e.g., neuromechanical simulations \cite{hase2003human}, when ethical or technical limitations prevent certain biological studies on live animals.

\pdfbookmark[2]{Limitation and future work}{Limitation and future work}
\subsection*{Limitation and future work}
To obtain general conclusions on state importance with minimal human bias and good statistical characteristics, we designed basic reward functions that encourage the exploration of feasible but slightly different motor policies across training sessions. Compared to the state-of-the-art locomotion \cite{peng2020learning,lee2020learning,yang2020multi}, we do not rely on reference trajectories or heavily fine-tuned reward terms to achieve natural-looking gaits. This allows the feedback control policy to fully exploit the solution space, drawing general conclusions about the importance of different states.

The identified key important states enabled successful learning of motor skills, demonstrating robustness to uncertainties in sensing, environment, and robot dynamics (Supplementary Video 1-4), which suggests that our saliency analysis based on the neural network, i.e., the mapping from filtered state input to unfiltered action output, has captured essential features of the overall policy for the identification of key feedback states. As future work, further investigation of other components of the framework and the properties of the overall policy would be an interesting direction. While considering the accuracy of sensory feedback, the importance ranking of feedback state can be further refined by composing the saliency map and sensitivity matrix of sensor noise levels, see details in Supplementary Note 4.

In future applications, in case certain states are identified as important but unreliable due to hardware limitations, one potential solution is to train a neural network to infer the estimation of such error-prone states from more reliable feedback. Prior work \cite{ji2022concurrent} has demonstrated the feasibility of such an approach, e.g., estimating body velocity and foot contact from joint positions measured by motor encoders, and the gravity vector plus base angular velocities obtained from IMU measurements.

\pdfbookmark[1]{Methods}{Methods}
\section*{Methods}
\pdfbookmark[2]{Robot platform}{Robot platform}
\subsection*{Robot platform}
We chose the \textit{A1} quadruped robot \cite{unitree} for our study, which is close to a real small-size dog and commonly used for locomotion research (see robot specifications in Supplementary Table \ref{tab:robot}). The extensive simulation validations were conducted in a physics-based simulation -- PyBullet \cite{coumans2019}. All the robot locomotion tasks were simulated in PyBullet with high-fidelity physics, and resulted robot motions and data were rendered in Unity \cite{haas2014history} for high resolutions snapshots and videos, which provide better visualization quality of the physical interactions and movements. 

\pdfbookmark[2]{Saliency analysis}{Saliency analysis}
\subsection*{Saliency analysis}\label{sec:saliency}
\pdfbookmark[3]{Integrated gradients}{Integrated gradients}
\subsubsection*{Integrated gradients}
Our use of the saliency analysis is inspired by feature attribution methods in image classification and explainable artificial intelligence \cite{simonyan2013deep,yosinski2015understanding,sundararajan2017axiomatic,jimenez2020drug}. There are several saliency methods or attribution methods, such as integrated gradients \cite{sundararajan2017axiomatic}, guided backpropagation \cite{springenberg2014striving}, DeepLift \cite{shrikumar2016not} and Grad-CAM \cite{selvaraju2017grad}. Here, we use \textit{integrated gradients} method to define the saliency values of feedback states for a learned policy. Integrated gradients method satisfies two axioms which are desirable for attribution methods \cite{sundararajan2017axiomatic}: (i) \textit{sensitivity}, i.e., the attribution should be non-zero if each input and baseline lead to different outputs; and (ii) \textit{implementation invariance}, i.e., the attributions should be the same for two networks if the outputs of both networks are identical for all the inputs, regardless of the detailed implementation. Some other common saliency methods are not able to satisfy both.

For example, vanilla gradients of the output with respect to the input and guided backpropagation break the axiom \textit{sensitivity} \cite{shrikumar2016not}, which will result in the gradients focusing on irrelevant features. Another common technique DeepLift breaks the axiom \textit{implementation invariance}, where results may differ for the networks with same functionalities but different implementation. In summary, the use of integrated gradients allows us to identify feedback states that are truly relevant. The same conclusion holds for each type of locomotion skill regardless of the implementation of deep neural networks, as long as the outputs of two implementations are the same for the identical inputs. 

It shall be noted that the integrated gradients method is applicable to the state-action mapping that is differentiable, meaning that it can analyze the influence of feedback states of motor skills which can be represented by differentiable machine learning models. However, it cannot be applied to underlying state-action policies that are nondifferentiable. At time step $t$, for the $i$-th dimension of feedback states $x \in \mathbb{R}^n$, integrated gradients $G(x_{i,t})$ are defined as follows:

\begin{equation}
    G(x_{i,t})=\sum_{j=1}^m\left|\frac{(x_{i,t}-\hat{x}_{i,t})}{p} \sum_{k=1}^{p}  \frac{\partial F_j(\hat{x_t}+k/p(x_t-\hat{x}_t))}{\partial x_{i,t}}\right|
\end{equation}

where $F(x_t) \in \mathbb{R}^m$ are the generated actions at time step $t$, $\hat{x}_{i,t}$ is the baseline input zero, $p=25$ is the number of steps in the Riemann approximation of the integral. The partial derivative is computed through backpropagation by calling \texttt{tf.gradients()} in TensorFlow \cite{abadi2016tensorflow}. For a better visualization to reveal the relative importance among feedback states via saliency maps, we define the raw saliency value $S_d(x_{i,t})$ as follows instead of directly using the computed integrated gradients:

\begin{equation}
    \epsilon=\frac{1}{nN}\sum_{t=1}^{N}\sum_{i=1}^{n}G(x_{i,t})
\end{equation}

\begin{equation}
    S_d(x_{i,t})=\begin{cases}
		G(x_{i,t})-\epsilon, & G(x_{i,t})>\epsilon\\
		0, &\mathrm{else}
		\end{cases}
\end{equation}

where, $N$ is the number of total time steps during the entire motion. The raw saliency value $S_d(x_{i,t})$ is further normalized to the range of $[0,1]$ as follows:

\begin{equation}
    S(x_{i,t})=S_d(x_{i,t})/\argmax_{{\substack{i \in \{1,2,...,n\} \\ t \in \{1,2,...,N\}}} } S_d(x_{i,t})
\end{equation}

\pdfbookmark[3]{Relative importance of feedback states}{Relative importance of feedback states}
\subsubsection*{Relative importance of feedback states}
For the $i$-th dimension of feedback states $x \in \mathbb{R}^n$, overall importance during the entire motion $I_i$ is computed as follows: 

\begin{equation}
    I_i=\sum_{t=1}^{N} S(x_{i,t})
\end{equation}

where $S(x_{i,t})$ is the saliency value for $x_i$ at time step $t$, and $N$ is the number of total time steps during the entire motion.  

For feedback state $o \in \mathbb{R}^h, (h \leq n)$, overall importance $I_o$ is computed as follows:

\begin{equation}
    I_o=\frac{1}{h} \sum_{q=1}^{h} I_{i(o,q)}
\end{equation}

where $i(o,q)$ is the index of the dimension of feedback states $x$ which maps to the $q$-th dimension of feedback state $o$ (see Supplementary Table \ref{tab:dimension}).

This work considers nine feedback states in total. For feedback state $o$, relative importance $r_o$ is defined as follows:

\begin{equation}
   r_o=\frac{I_o}{\sum_{o=1}^9 I_o} 
\end{equation}

\pdfbookmark[2]{Feedback states for learning locomotion skills}{Feedback states for learning locomotion skills}
\subsection*{Feedback states for learning locomotion skills}
Here we introduce a longlist of nine candidate states used in legged locomotion and the way to measure or estimate them on a real robot:
(i) base position in the world frame which can be estimated using visual inertial odometry \cite{mourikis2007multi}, the three-dimensional base position was used rather than the base height alone for a fair comparison with other states;
(ii) normalized gravity vector in the robot's local frame which reflects the body orientation of the robot and can be computed using roll and pitch angle measurements from IMU; 
(iii) base angular velocity measured by IMU;
(iv) base linear velocity in the robot heading frame estimated by fusing leg kinematics and the acceleration from IMU; 
(v) joint position measured by motor encoders;  
(vi) joint velocity which is measured by motor encoders and further normalized by maximum joint velocity; 
(vii) joint torque which is measured by torque sensors and further normalized by maximum joint torque;
(viii) foot position relative to base in the robot heading frame which can be computed through forward kinematics;
and (ix) foot contact with the ground which is computed by applying sigmoid function to the L2 norm of contact force $F_i$ measured by the force sensor at the end of the $i$-th foot as follows:

\begin{equation}
\frac{1}{1+e^{-c_1(F_i-c_2)}}, i=1,2,3,4    
\end{equation}

where $c_1=c_2=2.0$. Thus, foot contact is continuous within the range of $[0,1]$ (an example of continuous foot contact is in Extended Data Figure \ref{ex:phase2}) without a discontinuous switch between zero and one as in a threshold function which may affect the differentiability of the neural network for applying our analysis. For learning periodic locomotion tasks, such as trotting and bounding, we included a two-dimensional feedforward phase vector $[sin(2\pi\phi),cos(2\pi\phi)]$ on top of the above set of feedback states to represent continuous temporal information that encodes phase $\phi$ from 0\% to 100\% of a gait period. At each time step, the phase increases by a constant increment without any phase resetting mechanisms and is computed as:

\begin{equation}
    \phi = \frac{k\mod (T f_c)}{T f_c}
\end{equation}

where, $k$ is the control step counter, $T$ is desired gait period, $f_c$ is the control frequency. The phase vector is the feedforward term and thus was excluded for the quantification and comparison of the state importance, since the focus of this study is on the feedback terms.

For full-state policies, this complete set of feedback states were used for balance recovery (without the feedforward phase vector), trotting and bounding. For key-state policies, the states used were different for the three locomotion skills (see Fig. \ref{fig:general}\capC). Specifically, states (ii) and (v) were used for balance recovery, states (ii), (iv), (v) and phase vector were used for trotting, states (ii), (iii), (iv), (v) and phase vector were used for bounding. Learning pacing and galloping skills used the same set of states as for bounding.

\pdfbookmark[2]{Key-pose taxonomy for effective exploration and learning}{Key-pose taxonomy for effective exploration and learning}
\subsection*{Key-pose taxonomy for effective exploration and learning}\label{sec:taxonomy}

During the training of control policies using full feedback states, the robot pose is randomly initialized at each training episode to encourage the exploration of diverse states, which is a technique commonly used in robot learning \cite{cite:hwangbo2019learningAgile,reda2020learning}. However, random initialization is not data efficient in exploring the state space for a type of locomotion task, as most robot configurations are a priori invalid, due to the physical feasibility of the balance criteria. In other words, most of the robot configurations are not balanced or very far away from the desired locomotion, and therefore the collected samples are skewed by invalid exploration and less efficient for learning. To this end, on top of the key feedback states, we propose key-pose taxonomy to initialize the robot configuration at each training episode, as seeding conditions to enable more effective exploration and learning. 

Inspired by animal locomotion \cite{alexander2013principles,biewener2018animal} and whole-body support pose taxonomy from humanoid robotics \cite{cite:borras2015wholeBody,cite:borras2017wholeBody}, given a specific locomotion task, we can design key-pose taxonomy which consists of representative robot-ground contact configurations and distinct robot poses. The robot-ground contact configurations are straightforward to obtain, since quadrupedal locomotion is well studied in biology and we can easily obtain representative contact phases, e.g., trotting of dogs and horses. Compared to the existing pose taxonomy \cite{cite:borras2015wholeBody,cite:borras2017wholeBody} which is for the classification and inter-transitions of loco-manipulation, our proposed key-pose taxonomy here aims at task-specific effective learning.

Specifically, we use the configuration space of a floating-based robot to define each key pose, which is composed of body height, body orientation (roll and pitch angles), and joint angles. The base linear velocity and base angular velocity were set as zero at the start of each episode. Given the same contact configuration, we shall note that a quadruped robot may have multiple poses. For example, crouching and standing share the same robot-ground contact by four feet. Therefore, to balance the aspect of diversity, based on each ground contact or gait phase, we can use the robot configuration space to define multiple distinct key poses, so as to increase the number of initial poses which can sparsely cover the feasible motions related to a task. 

Following this principle, we designed five key poses for balance recovery, six for trotting, four for bounding, four for pacing, and five for galloping as shown in Fig. \ref{fig:flowchart}\capB ~and Supplementary Figure \ref{taxonomy}. The key poses within the designed taxonomy were sampled to initialize the robot pose at each training episode for each task, and the detailed transitions between the key poses will be explored during the learning process and thus obtained as a natural outcome. Using key-pose taxonomy as initial posture setting makes learning more efficient by narrowing the solution space for learning compared to random exploration.

In this work, we assigned the key poses within the taxonomy with equal probability for the DRL agent to encounter and explore upon. It shall be noted that the probability of each key pose can be more flexible. For example, we can assign higher probabilities to the key poses which are task-relevant but less likely to be encountered in natural interactions with the environment.

\pdfbookmark[2]{Quantitative metrics of performance evaluation}{Quantitative metrics of performance evaluation}
\subsection*{Quantitative metrics of performance evaluation}\label{score}
To quantify the performance for comprehensive comparison, a set of performance metrics, $S$, is designed for each task. For balance recovery, the performance metric set $S_{recovery}=\{s_{\tau},s_r, s_f,s_{h_N},s_{\phi_N}\}$. For trotting and bounding, the performance metric sets $S_{trotting}$ and $S_{bounding}$ are the same, i.e., $S=\{s_{\tau},s_v, s_{\psi},s_{h},s_{\phi}\}$. Note that the performance metrics are used for post-learning performance evaluations, which are not the same as the reward terms designed for learning in terms of formulations and weights for each physical quantity. 

The metric value for each physical quantity is in the range of $[0,1]$, and $N$ is the number of total time steps of an episode. Joint torque is used for the performance evaluation across all the three locomotion tasks. The performance metrics for this physical quantity are evaluated as follows. A higher value of joint torque metric indicates a more energy-efficient motion.

\begin{equation}
    s_{\tau}=1-\frac{1}{12N}\sum_{i=1}^{12} \sum_{t=1}^{N}|\tau_{i,t}|/\hat{\tau}
\end{equation}

where $\tau_{i,t}$ is the joint torque of the $i$-th joint at time step $t$, and $\hat{\tau}=33.5\;Nm$ is the maximum joint torque.

\pdfbookmark[3]{Performance metrics for balance recovery}{Performance metrics for balance recovery}
\subsubsection*{Performance metrics for balance recovery}
For balance recovery, the performance metrics for the other four physical quantities are evaluated as follows: 

(i) Recovery speed metric 

\begin{equation}
    s_r=1-T/\hat{T}
\end{equation}

where $T$ is the time duration of recovery to a standing posture and $\hat{T}$ is the time duration from the start of recovery to the end of an episode. A higher value indicates the recovery is completed within a shorter time period.

(ii) Final foot placement metric

\begin{equation}
    s_f=1-\frac{1}{8}\sum_{i=1}^8 |p_{f,i}-\hat{p}_{f,i}|/\hat{d}
\end{equation}

where $\mathbf{p}_{f}$ and $\hat{\mathbf{p}}_{f}$ are vectors of final and nominal foot positions in the horizontal plane of the robot heading frame. 
$\hat{\mathbf{p}}_{f}=[0.18m, 0.13m, -0.18m, -0.13m, -0.18m, 0.13m, 0.18m, -0.13m]$ and $\hat{d}=0.3\;m$ for \textit{A1} quadruped robot. A higher value indicates that the four feet are closer to the nominal foot positions at the end of recovery. 

(iii) Final body height metric

\begin{equation}
    s_{h_N}=\mathrm{min}(h_{N},\hat{h})/\hat{h}
\end{equation}

where $h_{N}$ is the body height at the end of recovery and $\hat{h}=0.25\;m$ is the nominal standing height of the robot. A higher value means that the final body height is closer to the nominal standing height of the robot. 

(iv) Final body orientation metric

\begin{equation}
    s_{\phi_N}=(\mathbf{g_{N}\hat{g}}+1)/2
\end{equation}

where $\mathbf{g_{N}}$ is the gravity vector of the robot at the end of recovery, and $\mathbf{\hat{g}}=[0,0,-1]$ is the nominal gravity vector of the robot. A higher value indicates that the final body orientation is closer to the nominal body orientation, i.e., zero roll angle and pitch angle. 

\pdfbookmark[3]{Performance metrics for trotting and bounding}{Performance metrics for trotting and bounding}
\subsubsection*{Performance metrics for trotting and bounding}
For trotting and bounding, the other four physical quantities are the same and the performance metrics are evaluated as follows:

(i) Forward velocity metric

\begin{equation}
    s_{v}=\mathrm{min}(\frac{1}{N}\sum_{t=1}^N V_{t},\hat{V})/\hat{V}
\end{equation}

where $V_t$ is the forward velocity in the horizontal plane at time step $t$, $\hat{V}$ is the nominal forward velocity, and $\hat{V}=0.5\;m/s$ for trotting and $\hat{V}=1.0\;m/s$ for bounding. A higher value indicates a faster average forward velocity during the entire episode. It shall be noted that a lower value for bounding does not indicate worse learning performance. The reason is that we did not set a desired forward velocity strictly in the reward for training bounding as for trotting. We set the desired velocity for training bounding as 1.0m/s. However, we do not penalize velocity higher than 1.0m/s to encourage higher velocity if possible. As a result, the robot may learn bounding policies with different average forward velocities with the same training settings.

(ii) Heading accuracy metric

\begin{equation}
    s_{\psi}=(\frac{1}{N}\sum_{t=1}^{N} \frac{\mathbf{v_{h,t}\hat{v}_h}}{|\mathbf{v_{h,t}}| |\mathbf{\hat{v}_h}|}+1)/2
\end{equation}

where $\mathbf{v_{h,t}}$ is the velocity vector of the robot in the horizontal plane of the robot heading frame at time step $t$, $\mathbf{\hat{v}_{h}}$ is the nominal velocity vector in the horizontal plane, $\mathbf{\hat{v}_{h}}=[0.5\;m/s,0\;m/s]$ for trotting and $\mathbf{\hat{v}_{h}}=[1.0\;m/s,0\;m/s]$ for bounding, and $|\cdot|$ is the magnitude of the vector. A higher value indicates better tracking of the nominal heading during the entire episode.

(iii) Body height metric

\begin{equation}
    s_h=\frac{1}{N}\sum_{t=1}^N \mathrm{min}(h_{t},\hat{h})/\hat{h}
\end{equation}

where $h_{t}$ is the body height at time step $t$, and $\hat{h}=0.3\;m$ is the nominal height of the robot. A higher value means that the body height is closer to the nominal height during the entire episode. 

(iv) Body orientation metric

\begin{equation}
    s_{\phi}=(\frac{1}{N}\sum_{t=1}^{N} \mathbf{g_{t}\hat{g}}+1)/2
\end{equation}

where $\mathbf{g_{t}}$ is the gravity vector of the robot at time step $t$, and $\mathbf{\hat{g}}=[0,0,-1]$ is the nominal gravity vector of the robot. A higher value indicates that the body orientation is closer to the nominal body orientation during the entire episode, i.e., zero roll angle and pitch angle.

\pdfbookmark[3]{Metrics for task-performance evaluation}{Metrics for task-performance evaluation}
\subsubsection*{Metrics for task-performance evaluation}
Given the five individual performance metrics for each locomotion task, we can further evaluate overall performance of key-state policies with respect to full-state policies. Consider a task-related physical quantity $i$, metric for the key-state policy $s_{i,key}$ and metric for the full-state policy $s_{i,full}$, overall performance of the key-state policy $s_{key}$ is defined as follows:

\begin{equation}
    s_{key}=\frac{1}{5}\sum_{i=1}^{5}\frac{s_{i,key}}{s_{i,full}}.
\end{equation}

Additionally, we can compute the mean of $s_{key}$ for multiple tasks to statistically evaluate the overall performance (balance recovery, trotting, bounding) in a statistical manner. The overall performance of irrelevant-state policies and open-loop trajectories are evaluated in the same approach. 

\bibliographystyle{plainnat}
\bibliography{arxiv}

\pagebreak
\captionsetup[figure]{labelfont={bf},labelformat={default},labelsep=bar,name={Extended Data Fig.},font=captionfont,list=no}  
\setcounter{figure}{0} 
\setcounter{equation}{0} 
\setcounter{table}{0} 
\setcounter{page}{1} 

\begin{figure}[H]    
    \centering
        {\includegraphics[width=0.9\textwidth]{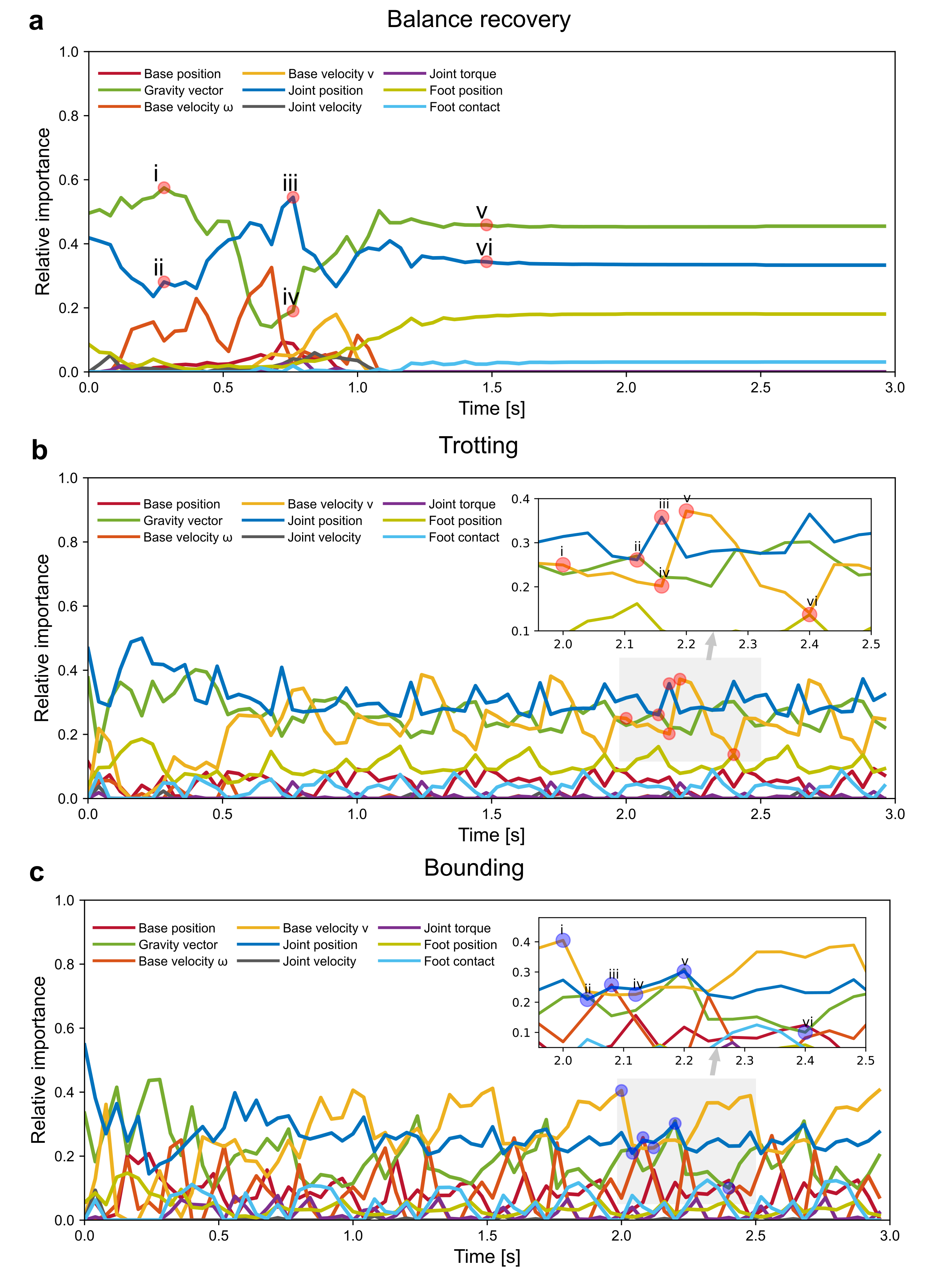}}
    \caption{\textbf{Relative importance of nine feedback states during $0-3s$.}\figcap{A} Balance recovery. \figcap{B} Trotting. \figcap{C} Bounding.}
    \label{ex:saliency-time}    
\end{figure}

\begin{figure}[H]    
    \centering
        {\includegraphics[width=0.9\textwidth]{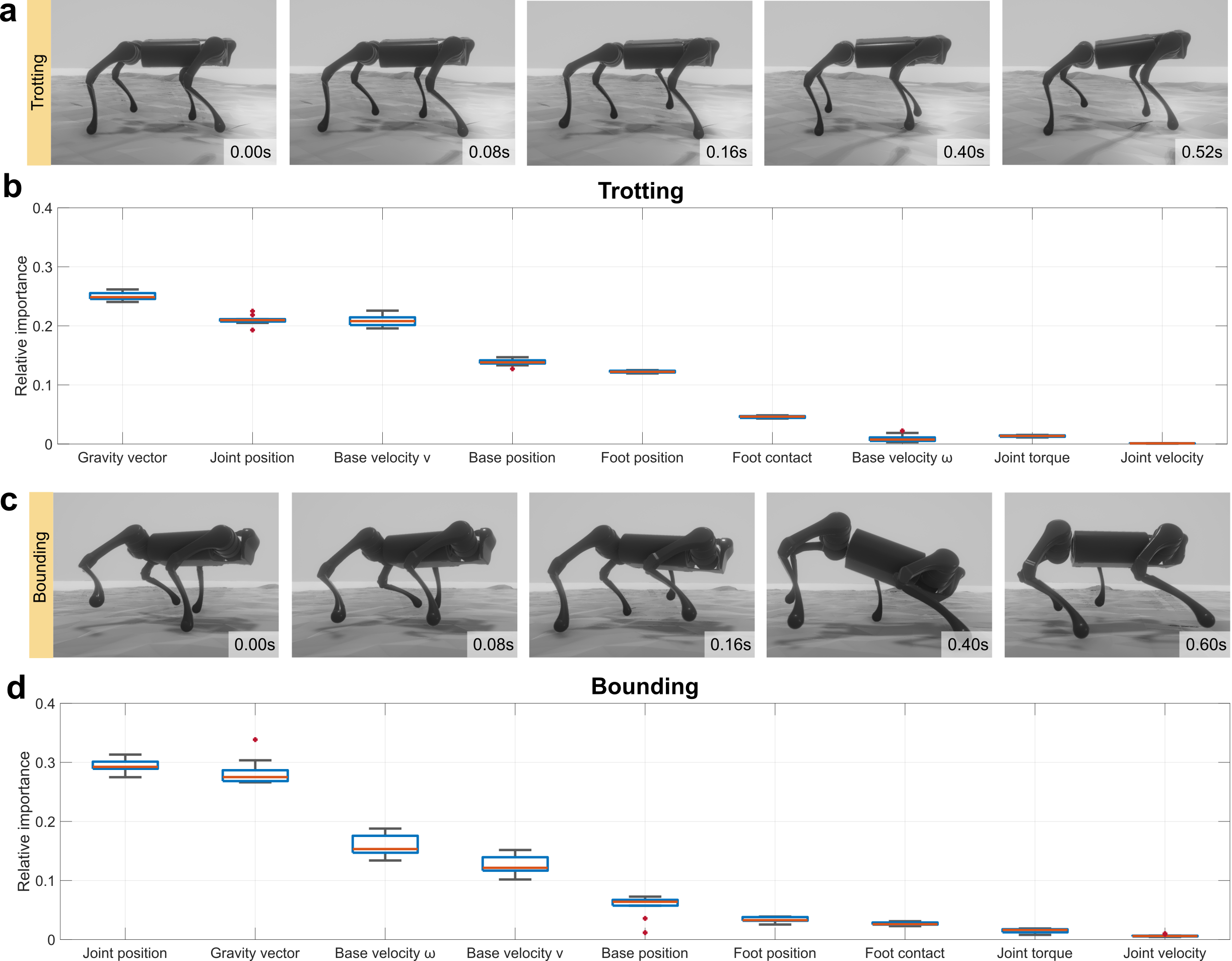}}
    \caption{\textbf{Key feedback states for locomotion on an uneven terrain with a maximum height of $3cm$ of the randomly generated irregular surfaces.} \figcap{A} The learned trotting over the rough terrain. \figcap{B} Boxplot showing the importance ranking of nine feedback states for trotting on the random terrain. \figcap{C} The learned trotting over the rough terrain. \figcap{D} Boxplot showing the importance ranking of nine feedback states for bounding on the random terrain.}
    \label{ex:boxplot-rough}    
\end{figure}

\begin{figure}[H]    
    \centering
        {\includegraphics[width=0.9\textwidth]{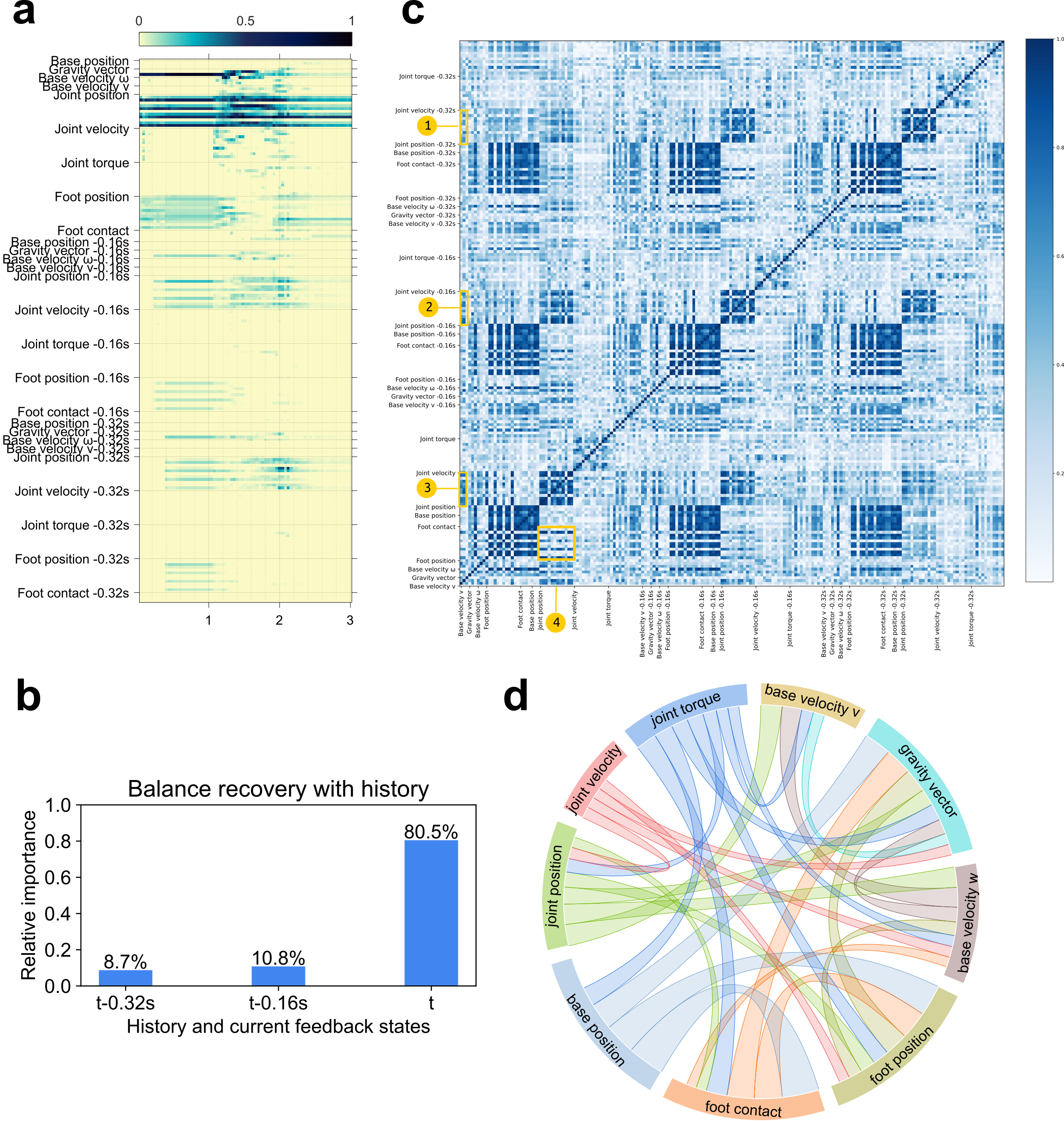}}
    \caption{\textbf{Analysis of the impact of history state information for the learned balance recovery policy.} \figcap{A} Saliency map showing the importance of feedback states at current time step and two history steps (0.16s and 0.32s ago) during $0-3s$. \figcap{B} Relative importance of all feedback states at the current timestep and two history steps. \figcap{C} Heatmap showing the absolute Pearson correlation coefficient between any two dimensions of feedback states including current and two history steps, where a darker color indicates a stronger correlation. Current body velocity has relatively high correlation with current joint positions (label 3) and two historical joint positions (label 1\&2), while the correlation becomes weaker as the history traces back. \figcap{D} Chord diagram showing the correlation between any two feedback states at current time step (self-correlation and percentage $<25\%$ are removed for clarity), where the wider the link between the two states, the stronger they correlate with each other.}
    \label{ex:history}    
\end{figure}

\begin{figure}[H]    
    \centering
        {\includegraphics[width=0.9\textwidth]{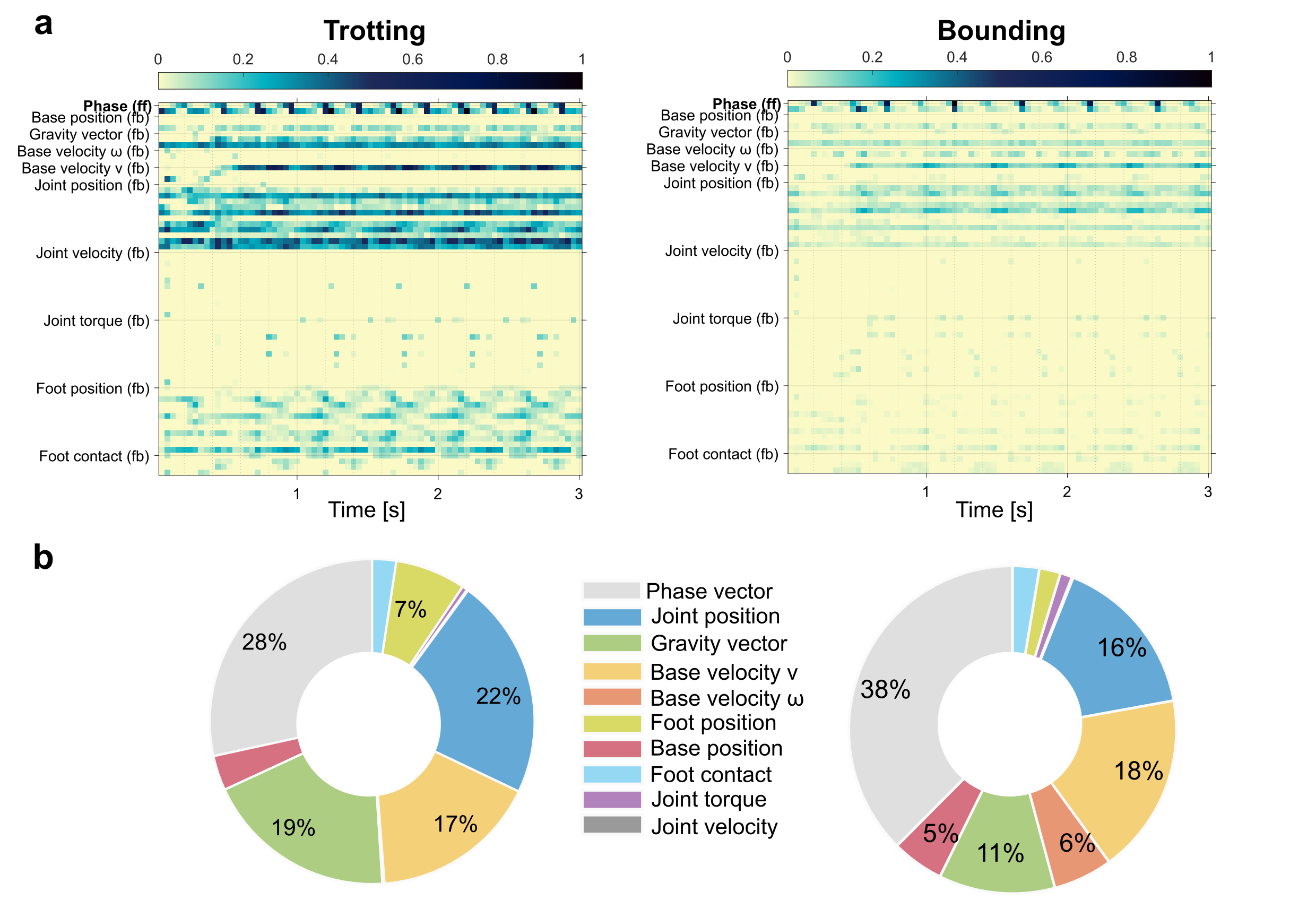}}
    \caption{\textbf{Analysis of the relative importance of the feedforward phase vector for trotting and bounding.} \figcap{A} Saliency maps showing the variation of the importance of the feedforward (\textbf{ff}) phase vector $(sin2\pi\phi, cos2\pi\phi)$ and the measured feedback states (\textbf{fb}). \figcap{B} Doughnut charts showing the relative importance of the feedforward (\textbf{ff}) phase vector (28\%, 38\%) and the feedback states (\textbf{fb}) (72\%, 62\%) for trotting and bounding, respectively.}
    \label{ex:phase1}    
\end{figure}

\begin{figure}[H]    
    \centering
        {\includegraphics[width=0.9\textwidth]{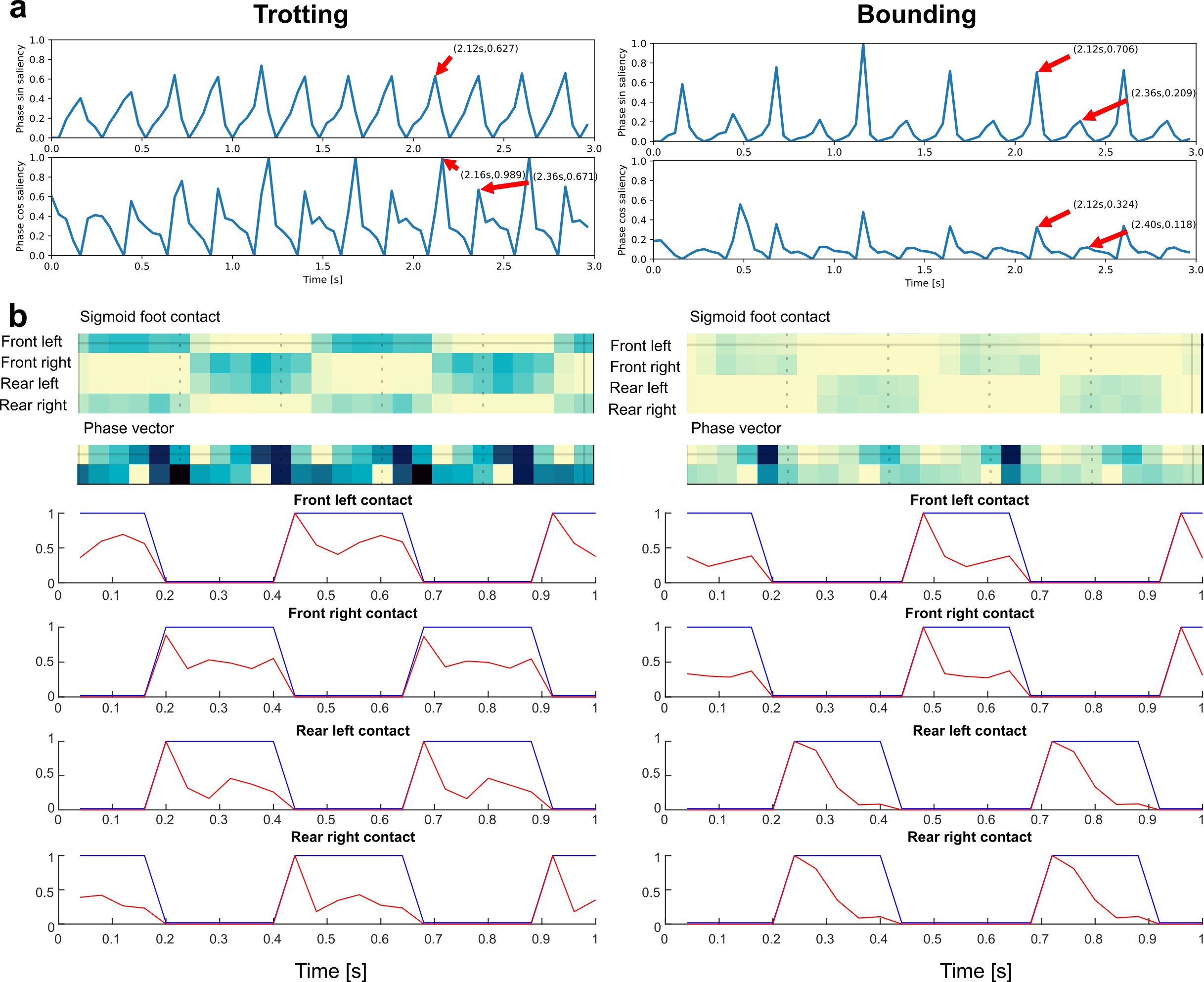}}
    \caption{\textbf{Analysis of the relative importance of the feedforward phase vector between swing and stance for trotting and bounding.} \figcap{A} Time plots of saliency values of the phase vector during $0-3 s$. \figcap{B} Saliency maps showing the variation of the phase vector importance between swing and stance and time plots of normalized foot contact feedback during $2-3s$ (two gait periods).}
    \label{ex:phase2}    
\end{figure}

\begin{figure}[H]    
    \centering
        {\includegraphics[width=0.9\textwidth]{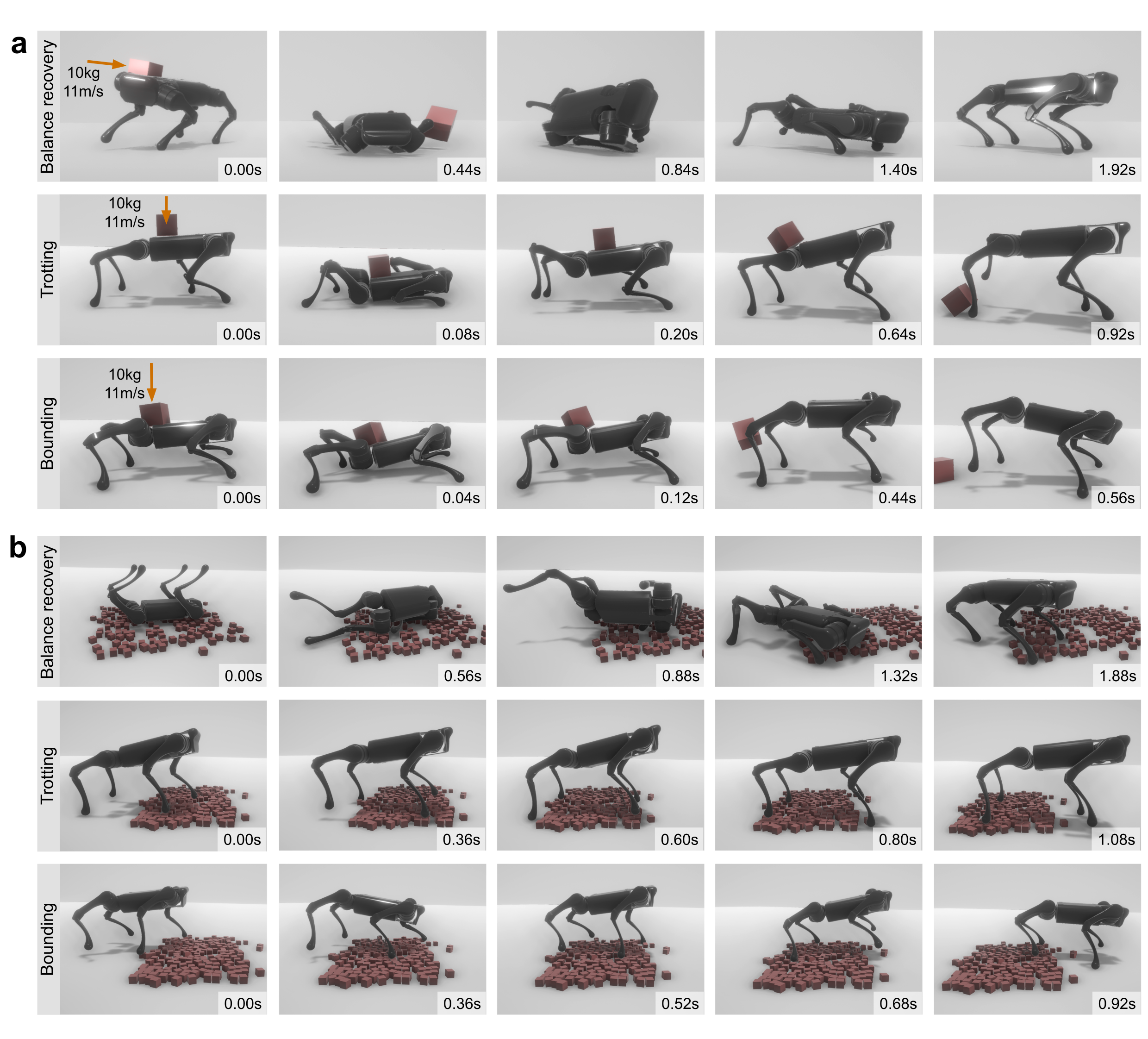}}
    \caption{\textbf{Robustness tests of key-state policies for balance recovery, trotting and bounding against unexpected perturbations.} \figcap{A} Perturbation by a $10 kg$ flying box at $11 m/s$ initial velocity for balance recovery (top), trotting (middle) and bounding (bottom). \figcap{B} Stable traversal over unseen rubble for balance recovery (top), trotting (middle) and bounding (bottom).}
    \label{ex:robust-key}    
\end{figure}

\pagebreak

\captionsetup[figure]{labelfont={bf},labelformat={default},labelsep=period,name={Supplementary Figure},list=yes}  
\setcounter{figure}{0} 
\setcounter{equation}{0} 
\setcounter{table}{0} 
\setcounter{page}{1} 
\pagebreak

\begin{center}
\pagebreak
\pdfbookmark[1]{Supplementary Information}{Supplementary Information}
\section*{\normalfont Supplementary Information for}
{\Large \textbf{Identifying Important Sensory Feedback for Learning Locomotion Skills}}

Wanming Yu$^{1,*}$, Chuanyu Yang$^{1,2,*}$, Christopher McGreavy$^{1}$, Eleftherios Triantafyllidis$^{1}$, Guillaume Bellegarda$^{3}$, Milad Shafiee$^{3}$, Auke J. Ijspeert$^{3}$, and Zhibin Li$^{4,\dagger}$

$^{1}$University of Edinburgh, Edinburgh, UK.

$^{2}$Shenzhen Amigaga Technology Co Ltd, Shenzhen, China.

$^{3}$École Polytechnique Fédérale de Lausanne (EPFL), Lausanne, Switzerland.

$^{4}$University College London, London, UK.

$^{*}$These authors contributed equally: Wanming Yu, Chuanyu Yang.

$^{\dagger}$Corresponding author: alex.li@ucl.ac.uk
\end{center}

\pdfbookmark[2]{Overview}{Overview}
\section*{Overview}
This document includes a list of supplementary figures, tables, and notes with more technical details.

\pdfbookmark[2]{Summary of key concepts and ideas}{Summary of key concepts and ideas}
\section*{Summary of key concepts and ideas}
In the following sub-section, present a set of questions and answers which capture the main concepts and ideas discussed in the paper. These concise summaries aim to help readers to form key takeaway messages and understand the core content of the research.

\textbf{How to select feedback states through the proposed saliency analysis of motor skills?}

We first define a comprehensive list of nine feedback states related to common physical quantities of a legged robot. Then, we train the desired motor skill using this list of feedback states. The obtained full-state policies are then used to perform the saliency analysis. See Section \textit{\nameref{sec:saliency}} in the main text for more details. Through the saliency analysis, we can obtain an importance ranking (Fig. \ref{fig:saliency}\capC) among all the feedback states and then select feedback states for learning motor skills accordingly.

\textbf{In which cases the saliency analysis can be used? Which cases are not applicable?}

The proposed saliency analysis can be applied to any differentiable state-action mapping. A typical case is a differentiable neural network and is not restricted to a specific neural network architecture or implementation. However, it is not applicable to nondifferentiable models since gradients need to be computed.

\textbf{Can saliency analysis be applied to traditional control methods?} 

The saliency analysis cannot be directly applied to traditional control methods since it can only be applied to differentiable models. However, after some processing of the trajectories generated by traditional control methods, saliency analysis can be applied. One feasible way is to collect the state-action pairs from traditional control methods and find a neural network to model the state-action mapping via supervised learning. Then, the saliency analysis can be applied to the learned model.

\textbf{Is it possible to identify in advance sensor specifications (e.g., sensor type and location, etc.) needed to generate a specific robot motion?}

Identifying sensor specifications, such as sensor type and location, for a specific robot motion is a challenging task without prior motor skills or application scenarios. While domain knowledge and human heuristics may be applied, it is hard to formulate appropriate quantitative analysis and the investigation of sensor specifications is out of the scope of this paper. Our approach requires an existing neural network policy in the first place and exploits the simulated motions to obtain a differentiable state-action mapping of the motor skills, in the form of a neural network, to determine appropriate types of sensory feedback information for a given motion task. Though this requires additional computational resources, it is necessary for effective identification and analysis of key feedback states in motor skill learning, similar to other computer-/simulation-aided design approaches.

\textbf{How to quantify and benchmark the performance of each motor skill?}

We define five performance metrics for each motor skill. See more details in Section \nameref{score} in the main text. The value of each metric is between zero and one, and a higher value indicates a better performance correspondingly. The average value of all the five performance metrics indicates the overall performance of the motor skill, which can be used for benchmarking.

\textbf{What are the differences in terms of performance of the learned policies using full feedback states and key feedback states?}

Using the performance metrics defined for each motor skill, we demonstrate that the learned policies using key feedback states can achieve comparable overall performance to those using full feedback states across different locomotion skills in various test scenarios. Here, the performance of the full-state policies are used as \textit{baselines} to benchmark performance from other policies. As shown in Fig. \ref{fig:benchmark}\capD, the learned policies using only the key feedback states can achieve 94.7\%, 99.1\% and 93.7\% of the task performance of those using full feedback states on a flat ground for balance recovery, trotting and bounding, respectively.

Moreover, the learned key-state policies can also guarantee similar level of performance in robustness tests as shown in Supplementary Figure \ref{robust-metric}. When the robot is perturbed by a flying box, the the overall performance (balance recovery, trotting, bounding) of the key-state policies achieves 101.1\%, 100.6\% and 97.4\% respectively, compared to the baseline. The key-state policies can achieve 96.8\%, 101.3\% and 96.1\% of the overall performance for these three locomotion skills over unseen rubble terrain.

\pagebreak
{\textbf{Supplementary Information}}
\vspace{-1.5cm}
\listoffigures

\vspace{-1.5cm}
\listoftables

\vspace{-0.25cm}
\hspace{0.65cm}Supplementary Note 1. Deep reinforcement learning framework.

\vspace{-0.25cm}
\hspace{0.65cm}Supplementary Note 2. Performance benchmark.

\vspace{-0.25cm}
\hspace{0.65cm}Supplementary Note 3. Robustness tests.

\vspace{-0.25cm}
\hspace{0.65cm}Supplementary Note 4. Evaluation of impact of sensing accuracy.

\vspace{-0.25cm}
\hspace{0.65cm}Supplementary Note 5. Comparison of relative importance using maximum saliency value.

\vspace{-0.25cm}

\pdfbookmark[2]{Supplementary figures}{Supplementary figures}
\section*{Supplementary figures}

\begin{figure}[H]
\includegraphics[width=0.9\textwidth]{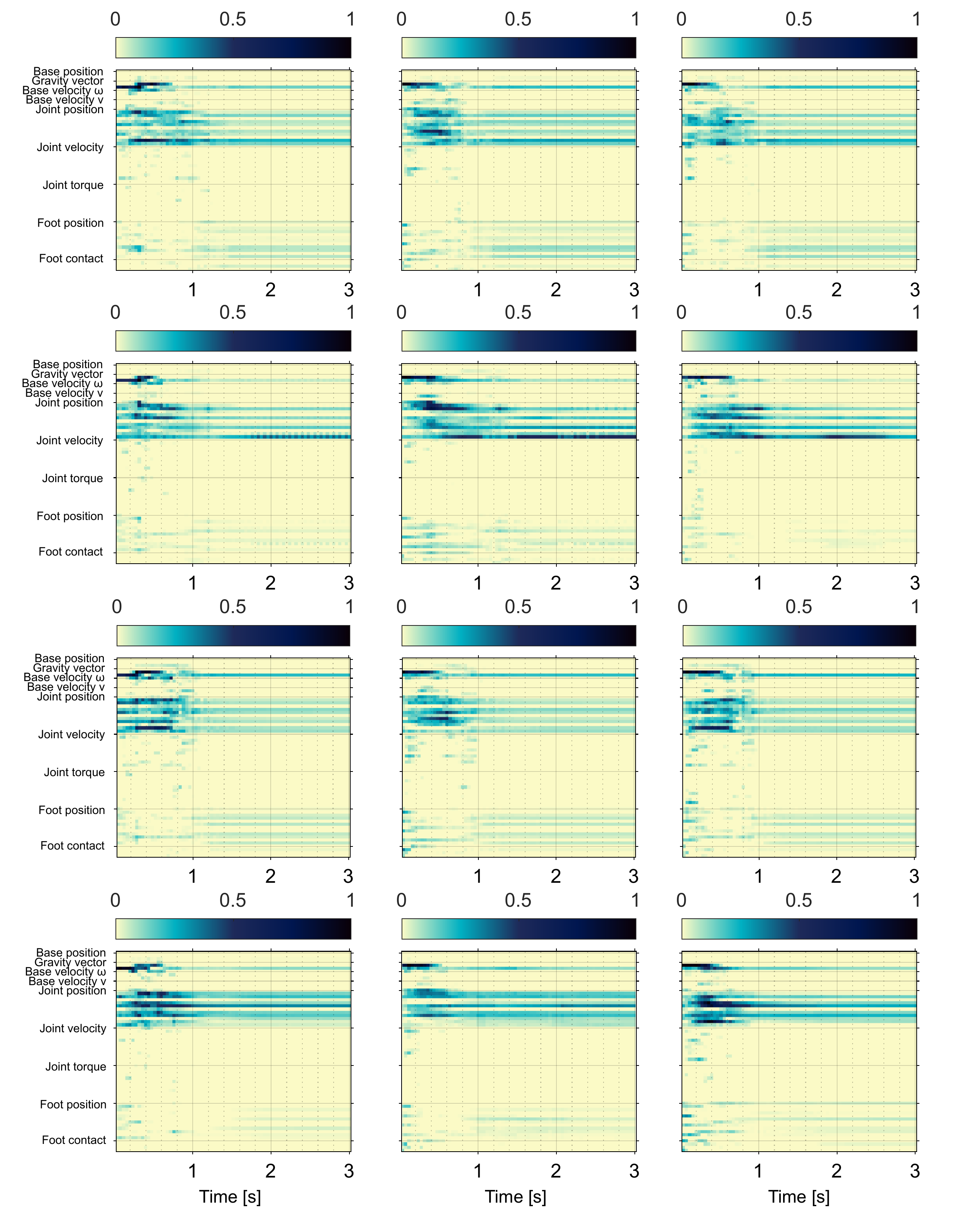}
\caption[Saliency maps generated by the saliency analysis of four different learned balance recovery policies.]{Saliency maps generated by the saliency analysis of four different learned balance recovery policies. Each row corresponds to one balance recovery policy executed with three different initial fall postures in the order of lying on the back, lying on the left and lying on the right. The saliency maps are for twelve different balance recovery motions in total.}
\label{saliency-stand}
\end{figure}

\begin{figure}[H]
\includegraphics[width=0.9\textwidth]{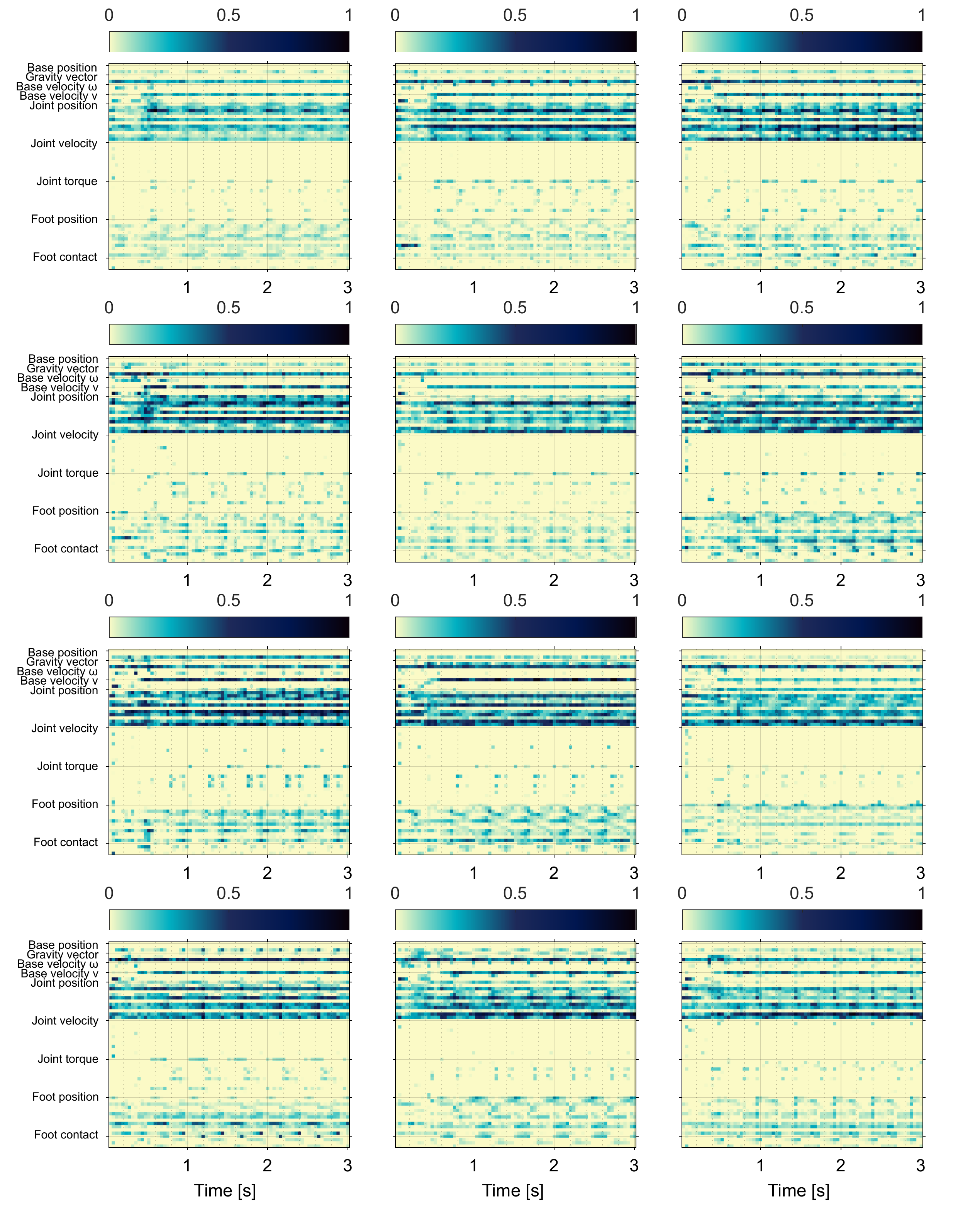}
\caption{Saliency maps generated by the saliency analysis of twelve different learned trotting policies.}
\label{saliency-trot}
\end{figure}

\begin{figure}[H]
\includegraphics[width=0.9\textwidth]{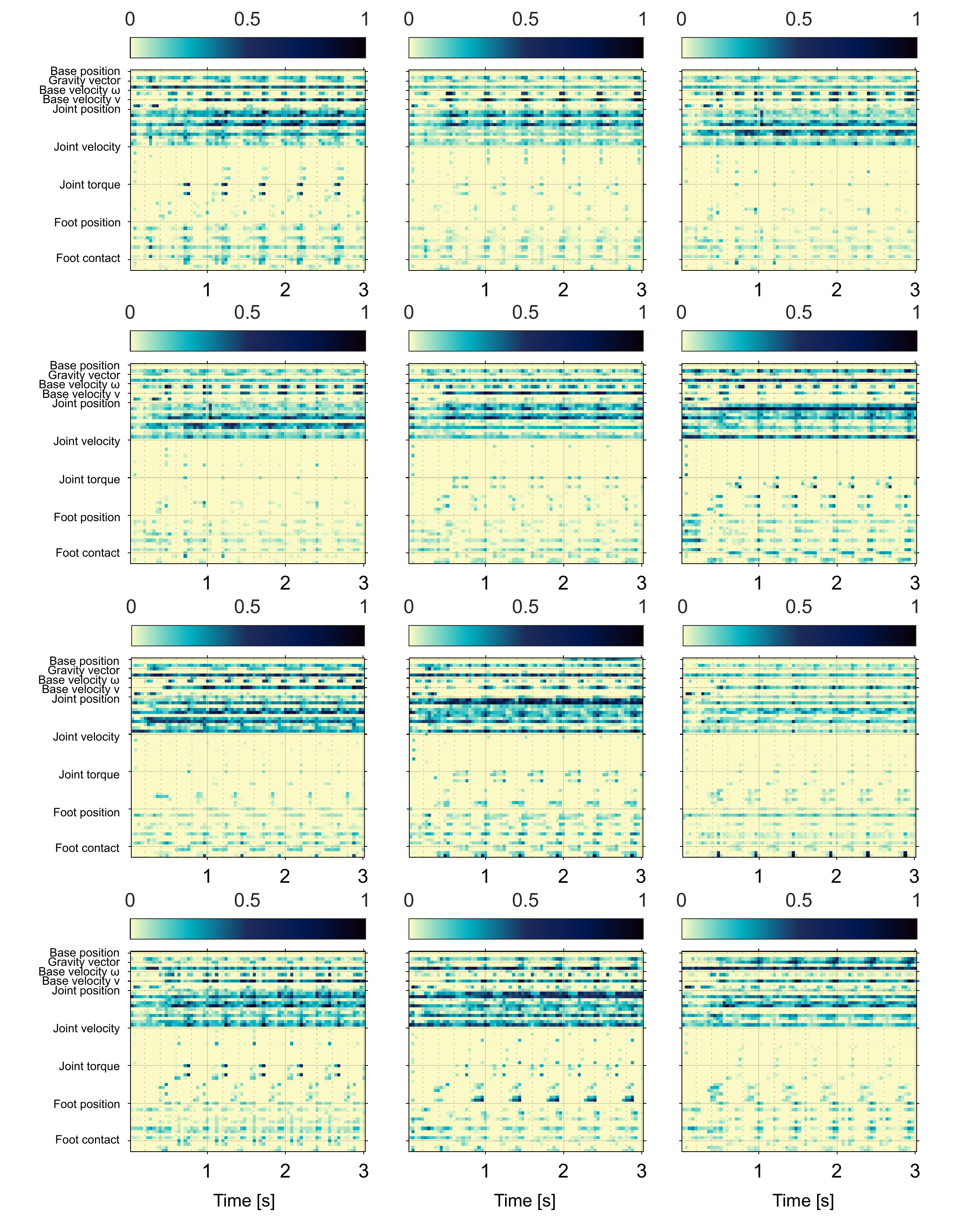}
\caption{Saliency maps generated by the saliency analysis of twelve different learned bounding policies.}
\label{saliency-bound}
\end{figure}

\begin{figure}[H]
\includegraphics[width=0.9\textwidth]{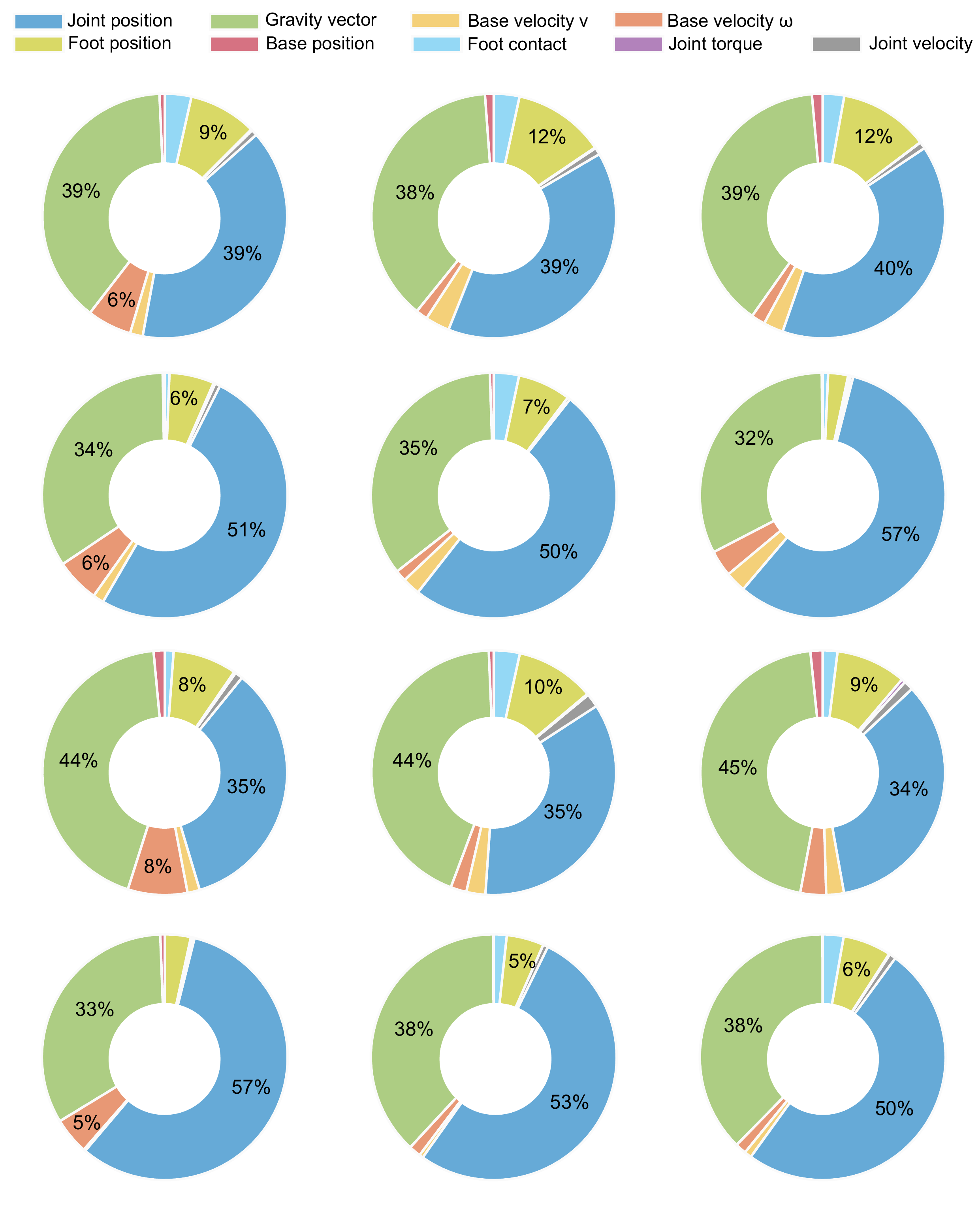}
\caption[Doughnut charts showing the relative importance of nine feedback states for four different learned balance recovery policies.]{Doughnut charts showing the relative importance of nine feedback states for four different learned balance recovery policies. Each row corresponds to one balance recovery policy executed with three different initial fall postures in the following order: lying on the back, lying on the left and lying on the right. Each doughnut chart shows the relative importance for a specific balance recovery motion, with twelve different motions in total (corresponding to the saliency maps in Supplementary Figure \ref{saliency-stand}).}
\label{pie-stand}
\end{figure}

\begin{figure}[H]
\includegraphics[width=0.9\textwidth]{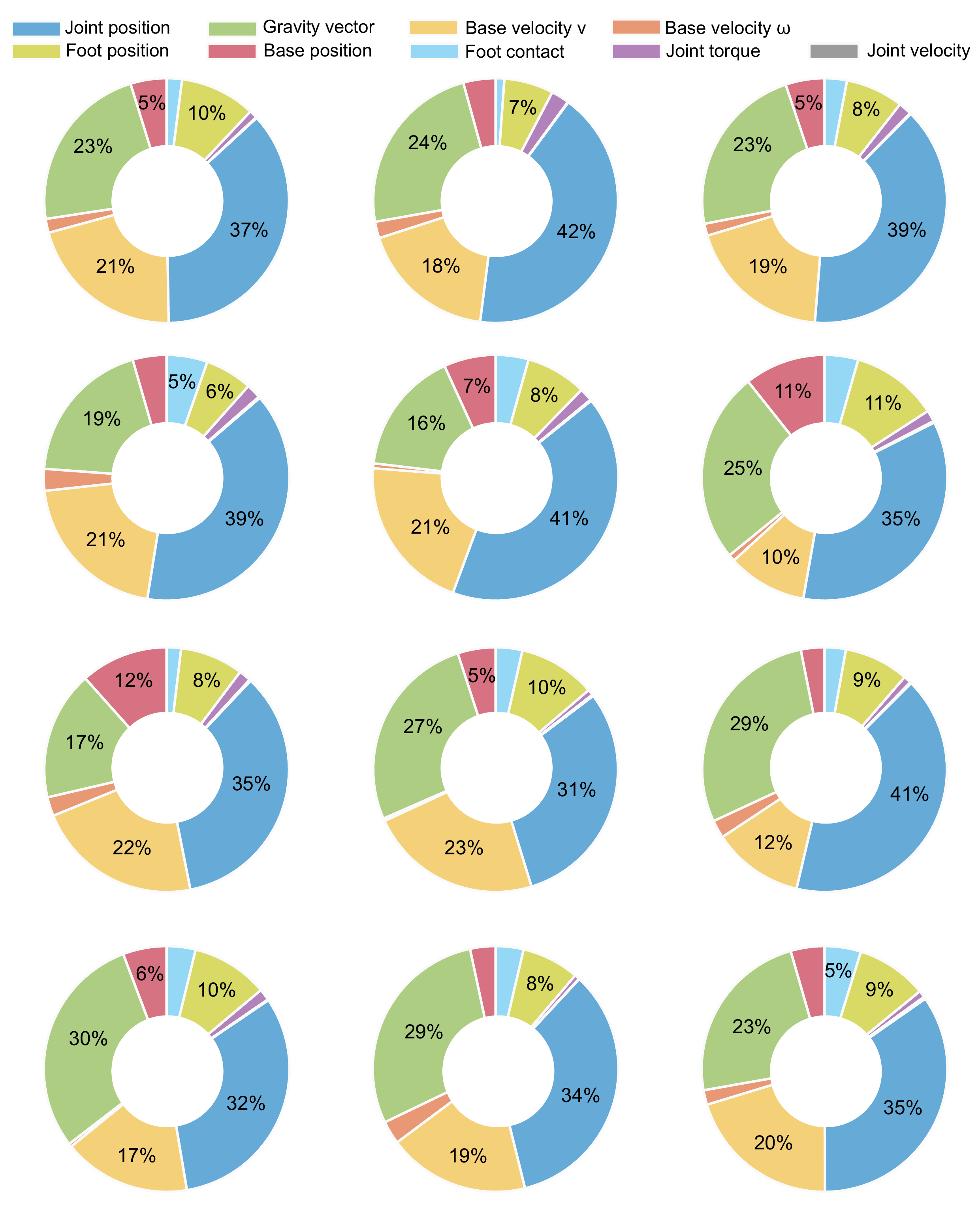}
\caption[Doughnut charts showing the relative importance of nine feedback states for twelve different learned trotting policies.]{Doughnut charts showing the relative importance of nine feedback states for twelve different learned trotting policies (corresponding to the saliency maps in Supplementary Figure \ref{saliency-trot}).}
\label{pie-trot}
\end{figure}

\begin{figure}[H]
\includegraphics[width=0.9\textwidth]{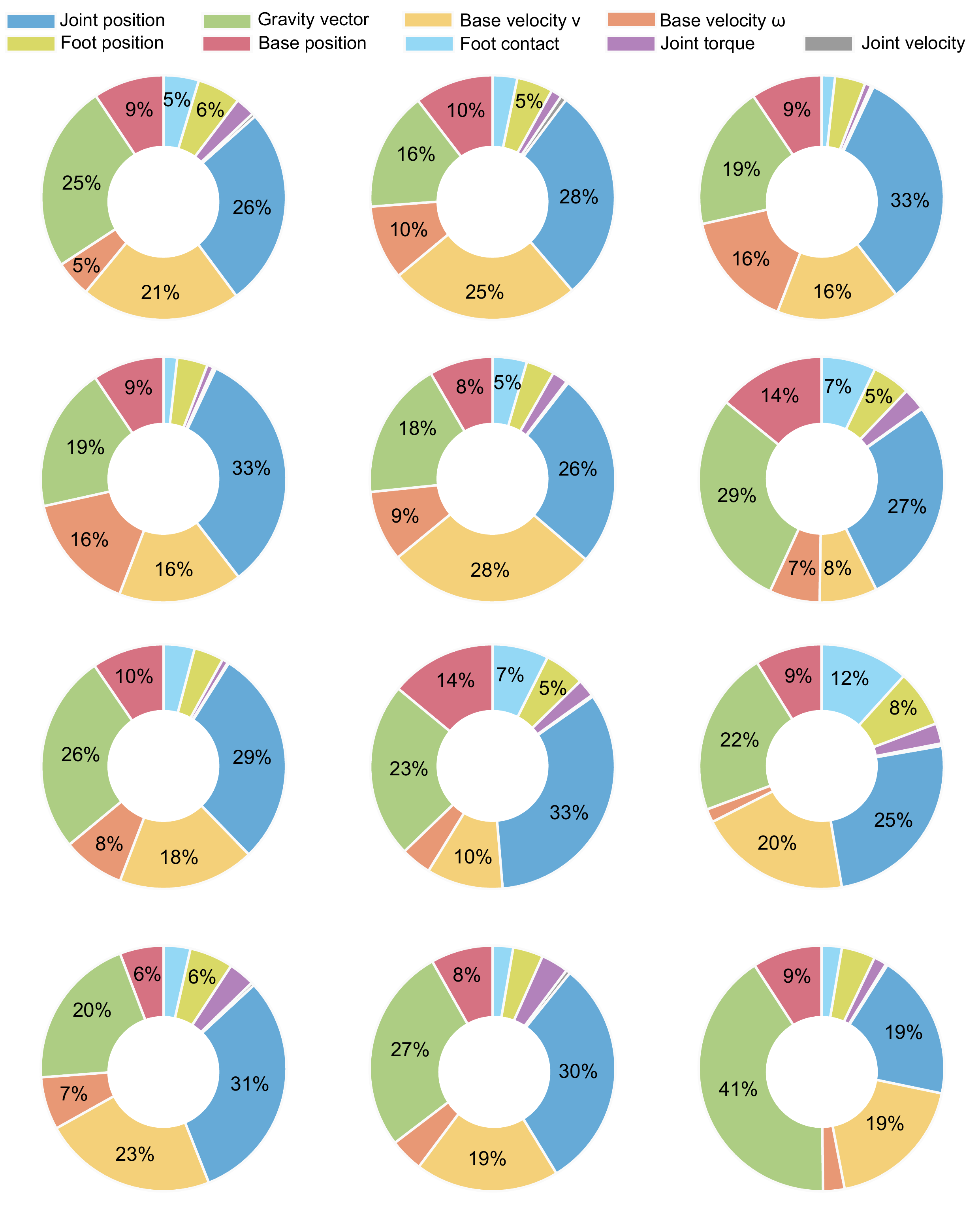}
\caption[Doughnut charts showing the relative importance of nine feedback states for twelve different learned bounding policies.]{Doughnut charts showing the relative importance of nine feedback states for twelve different learned bounding policies (corresponding to the saliency maps in Supplementary Figure \ref{saliency-bound}).}
\label{pie-bound}
\end{figure}

\begin{figure}[H]
\centering
\includegraphics[width=0.9\textwidth]{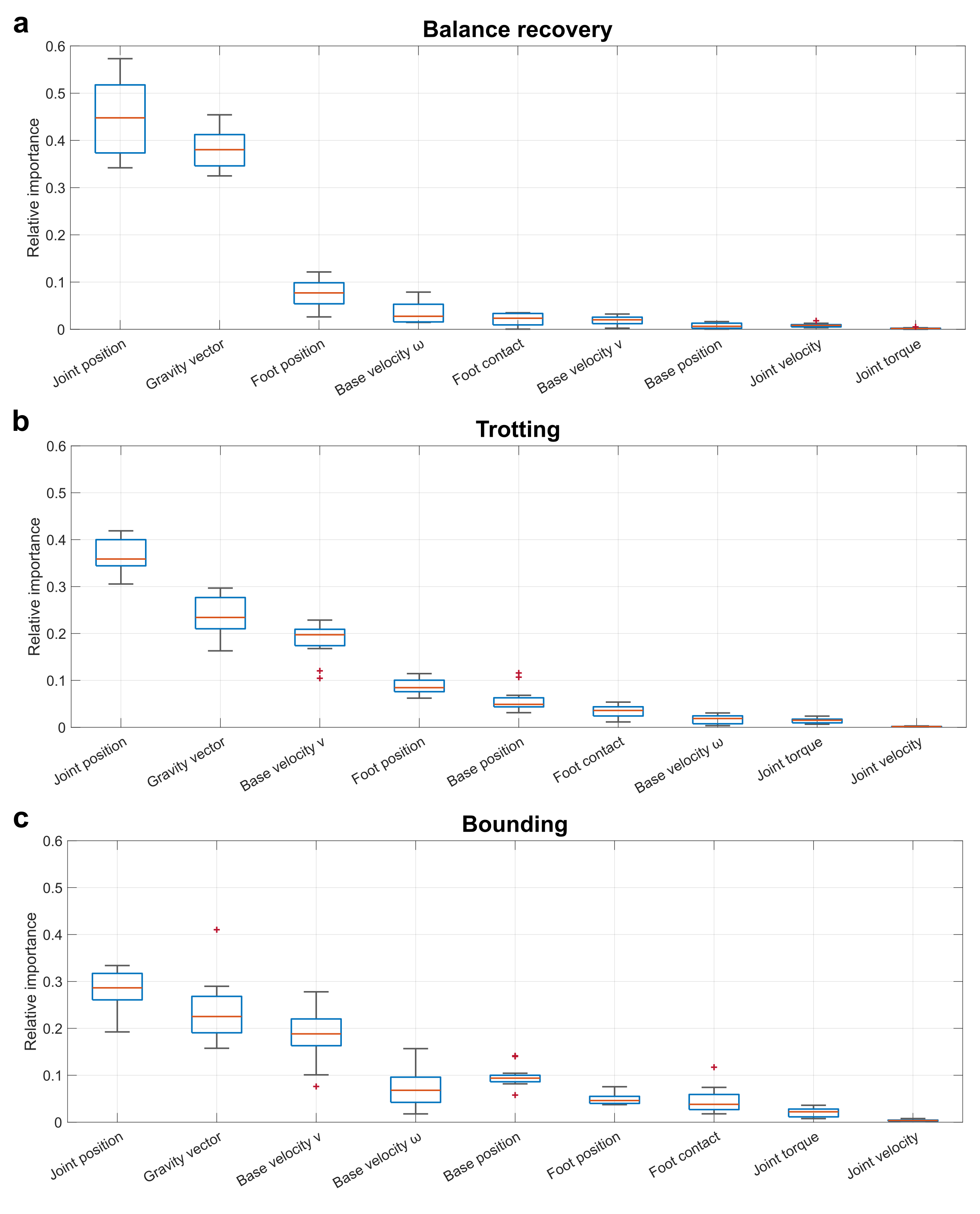}
\caption[Comparison of the relative importance of nine feedback states for balance recovery, trotting, and bounding, respectively.]{Comparison of the relative importance of nine feedback states for balance recovery, trotting, and bounding, respectively. Each box shows the median (red horizontal line), 25th and 75th percentiles (lower and upper blue horizontal lines), minimum and maximum (lower and upper gray horizontal lines), and outliers (red plus sign) of the relative importance of the corresponding feedback state from 12 different trials given one corresponding locomotion task. \figcap{A} A boxplot showing the relative importance of nine feedback states summarized from twelve different learned balance recovery motions, corresponding to Supplementary Figure \ref{saliency-stand} and \ref{pie-stand}. The key feedback states for balance recovery include joint position and gravity vector. \figcap{B} A boxplot showing the relative importance of nine feedback states summarized from twelve different learned trotting policies, corresponding to Supplementary Figure \ref{saliency-trot} and \ref{pie-trot}. The key feedback states for trotting include joint position, gravity vector, and base linear velocity.  \figcap{C} A boxplot showing the relative importance of nine feedback states summarized from twelve different learned bounding policies, corresponding to Supplementary Figure \ref{saliency-bound} and \ref{pie-bound}. The key feedback states for bounding include joint position, gravity vector, base linear velocity and base angular velocity. }
\label{saliency-boxplot}
\end{figure}

\begin{figure}[H]    
    \centering
        {\includegraphics[width=0.9\textwidth]{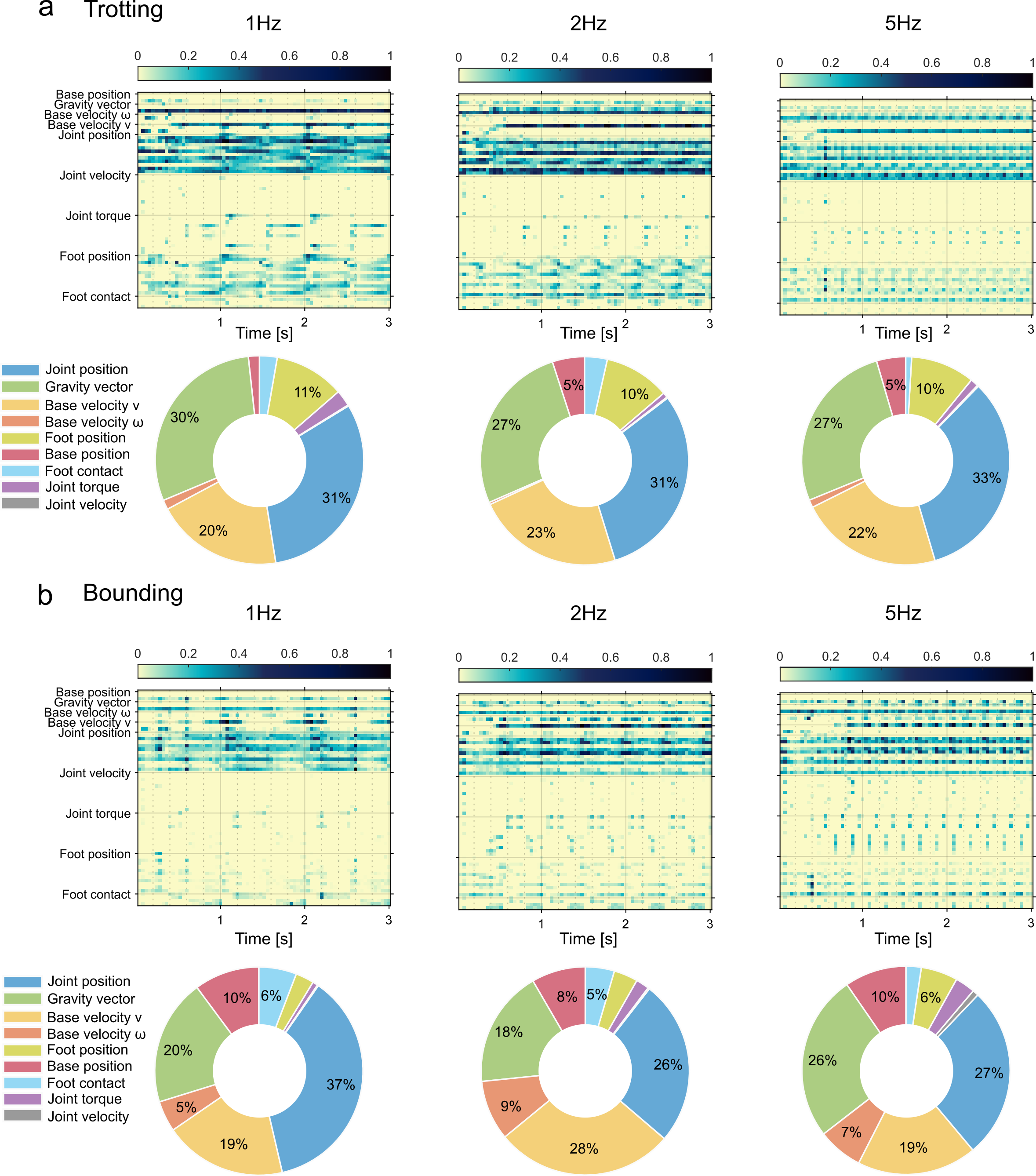}}
    \caption[Saliency analysis of locomotion gaits under low, medium and high frequencies.]{Saliency analysis of locomotion gaits under low, medium and high frequencies. \figcap{A} Saliency maps and doughnut charts for trotting under 1 Hz, 2 Hz and 5 Hz. \figcap{B} Saliency maps and doughnut charts for bounding under 1 Hz, 2 Hz and 5 Hz.}
    \label{frequency}    
\end{figure}

\begin{figure}[H]    
    \centering
        {\includegraphics[width=0.9\textwidth]{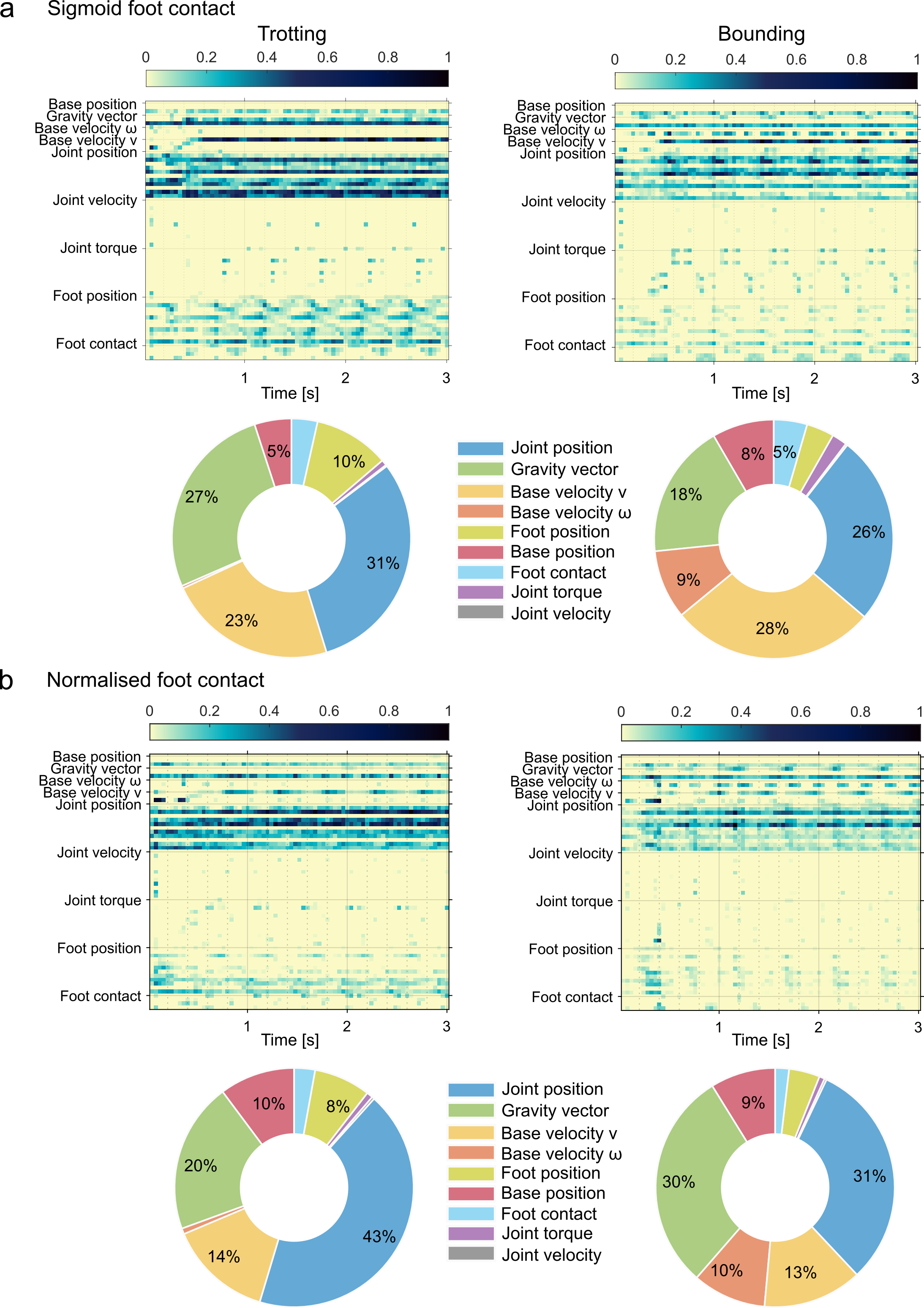}}
    \caption[Influence of different foot contact formulations on saliency analysis.]{Influence of different foot contact formulations on saliency analysis. \figcap{A} Saliency maps and doughnut charts for trotting and bounding using sigmoid foot contact. \figcap{B} Saliency maps and doughnut charts for trotting and bounding using normalised foot contact.}
    \label{contact}    
\end{figure}

\begin{figure}[H]    
    \centering
        {\includegraphics[width=0.9\textwidth]{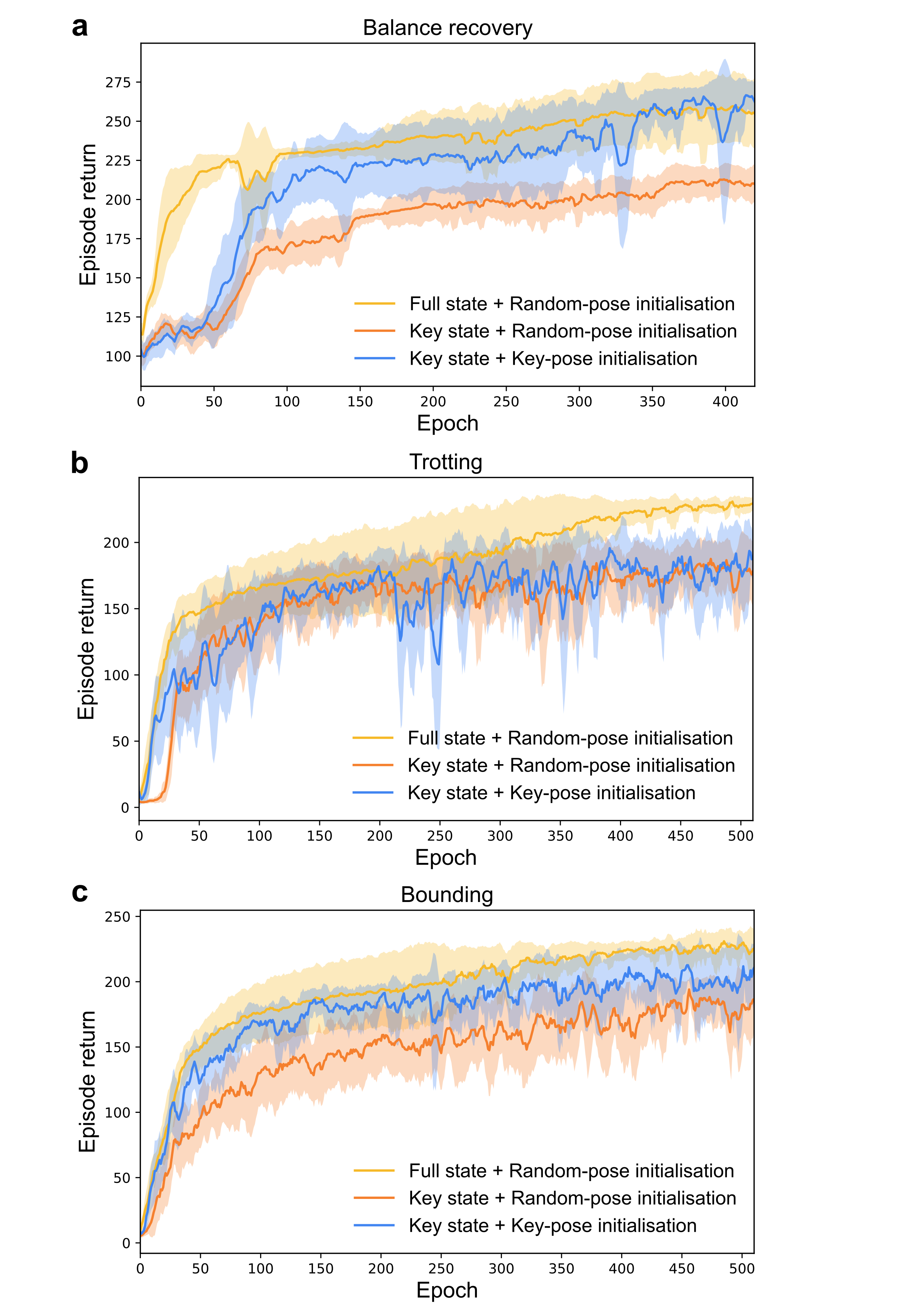}}
    \caption[Comparison of the learning curves for three cases for balance recovery, trotting and bounding, respectively.]{Comparison of the learning curves for three cases: full feedback states with random-pose initialization, key feedback states with random-pose initialization, and key feedback states with key-pose initialization, for balance recovery, trotting and bounding, respectively. For each learning curve, we first compute the mean and standard deviation of episode return from five learning sessions, and further apply moving average with a sliding window size of five epochs for better visualization. \figcap{A} Learning curves for balance recovery. \figcap{B} Learning curves for trotting. \figcap{C} Learning curves for bounding.}
    \label{learning-curve}    
\end{figure}

\begin{figure}[H]    
    \centering
        {\includegraphics[width=0.9\textwidth]{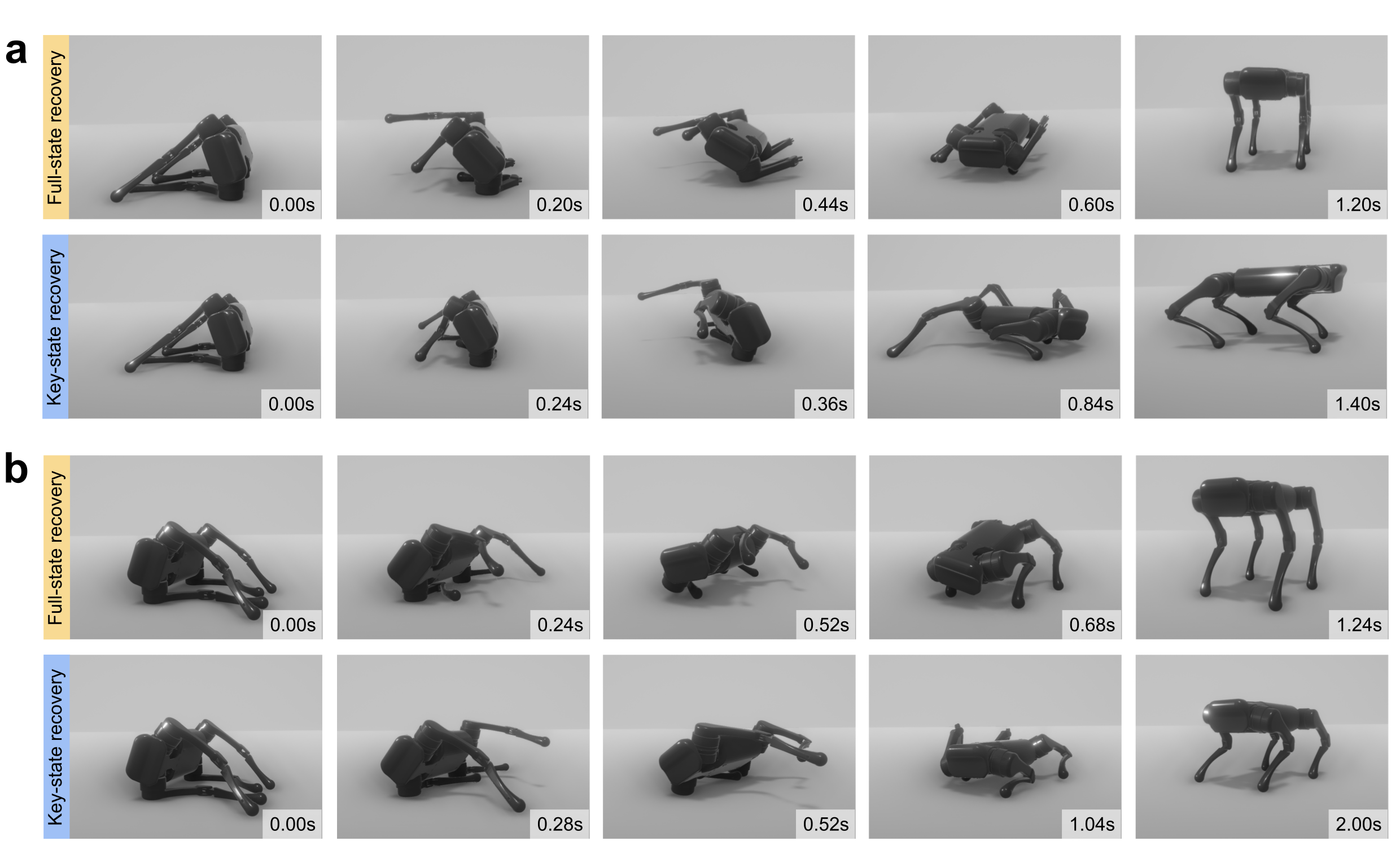}}
    \caption[Snapshots of \textit{A1} robot recovering from various initial fall postures.]{Snapshots of \textit{A1} robot recovering from various initial fall postures. \figcap{A} The robot recovers from lying on its left to a nominal standing posture using a full-state policy (top) and a key-state policy (bottom), respectively. \figcap{B} The robot recovers from lying on its right to a nominal standing posture using a full-state policy (top) and a key-state policy (bottom), respectively.}
    \label{recovery-snapshots}    
\end{figure}

\begin{figure}[H]    
    \centering
        {\includegraphics[width=0.9\textwidth]{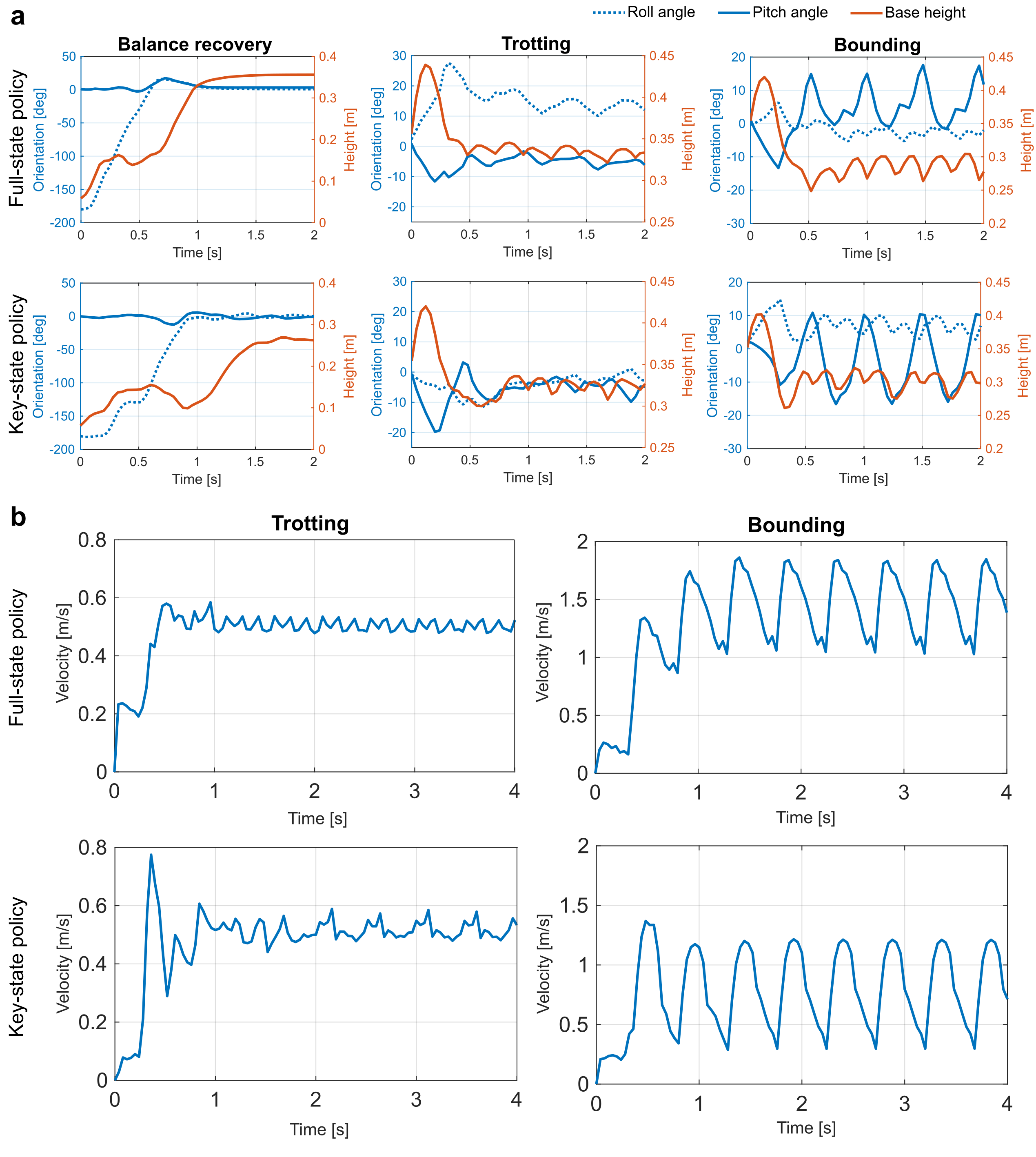}}
    \caption[Benchmarking of full-state policies and key-state policies for balance recovery, trotting and bounding.]{Benchmarking of full-state policies and key-state policies for balance recovery, trotting and bounding. Within each subfigure, the full-state policy is at the top and the key-state policy is at the bottom. \figcap{A} Time plots of body orientation (roll and pitch angles) and height for balance recovery, trotting and bounding, respectively. \figcap{B} Time plots of forward velocity for trotting and bounding, respectively.}
    \label{benchmark-detail}    
\end{figure}

\begin{figure}[H]    
    \centering
        {\includegraphics[width=0.9\textwidth]{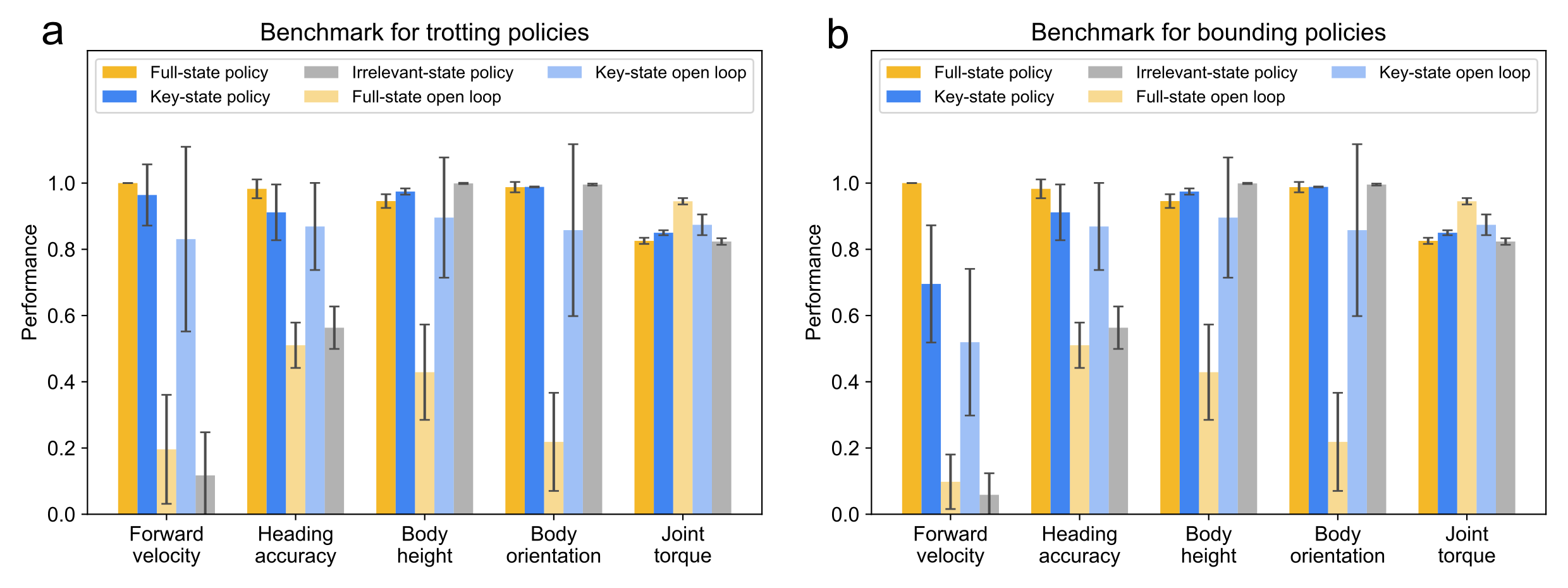}}
    \caption[Performance benchmark of full-state policies, key-state policies, irrelevant-state policies, full-state open-loop trajectories and key-state open-loop trajectories for training on flat ground.]{Performance benchmark of full-state policies, key-state policies, irrelevant-state policies, full-state open-loop trajectories and key-state open-loop trajectories for training on a flat ground. \figcap{A} Performance benchmark of trotting in ten scenarios. \figcap{B} Performance benchmark of bounding in ten scenarios.}
    \label{benchmark-std}    
\end{figure}

\begin{figure}[H]    
    \centering
        {\includegraphics[width=0.9\textwidth]{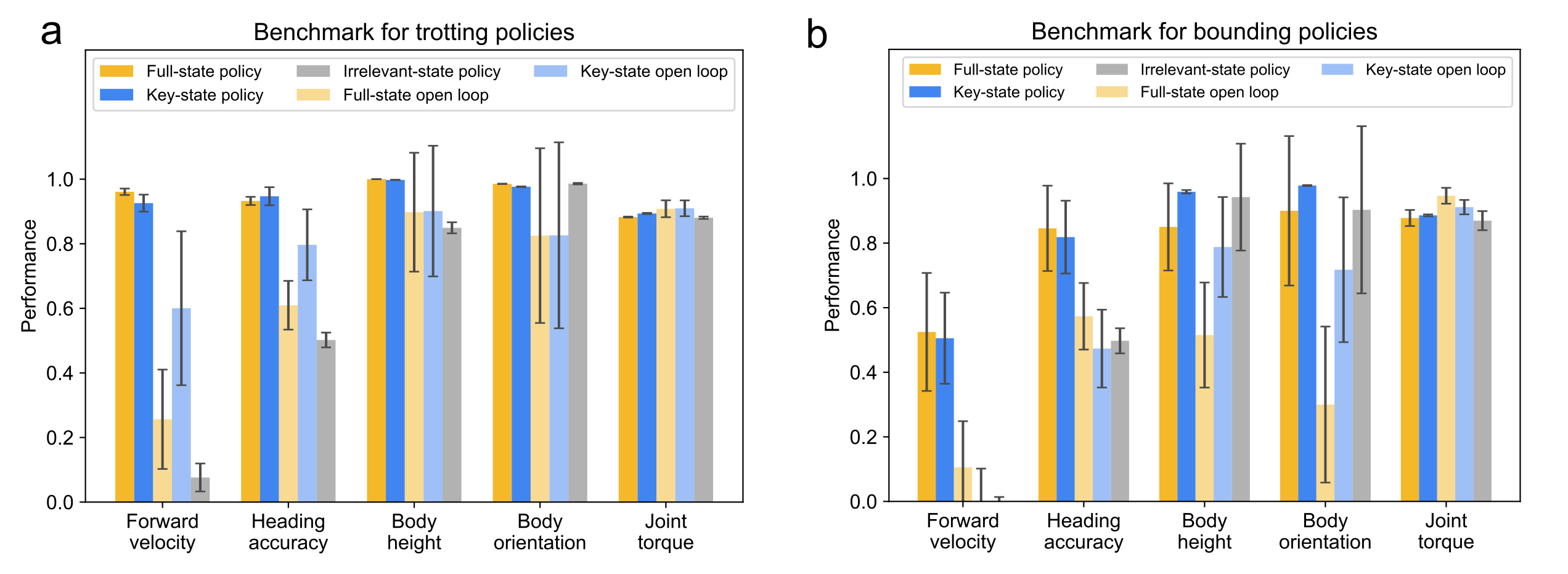}}
    \caption[Performance benchmark of full-state policies, key-state policies, irrelevant-state policies, full-state open-loop trajectories and key-state open-loop trajectories for training on a random terrain with a maximum height of 3 cm.]{Performance benchmark of full-state policies, key-state policies, irrelevant-state policies, full-state open-loop trajectories and key-state open-loop trajectories for training on a random terrain with a maximum height of 3 cm. \figcap{A} Performance benchmark of trotting on a rough terrain with ten random seeds. \figcap{B} Performance benchmark of bounding on a rough terrain with ten random seeds.}
    \label{benchmark-std-3cm}    
\end{figure}

\begin{figure}[H]    
    \centering
        {\includegraphics[width=0.9\textwidth]{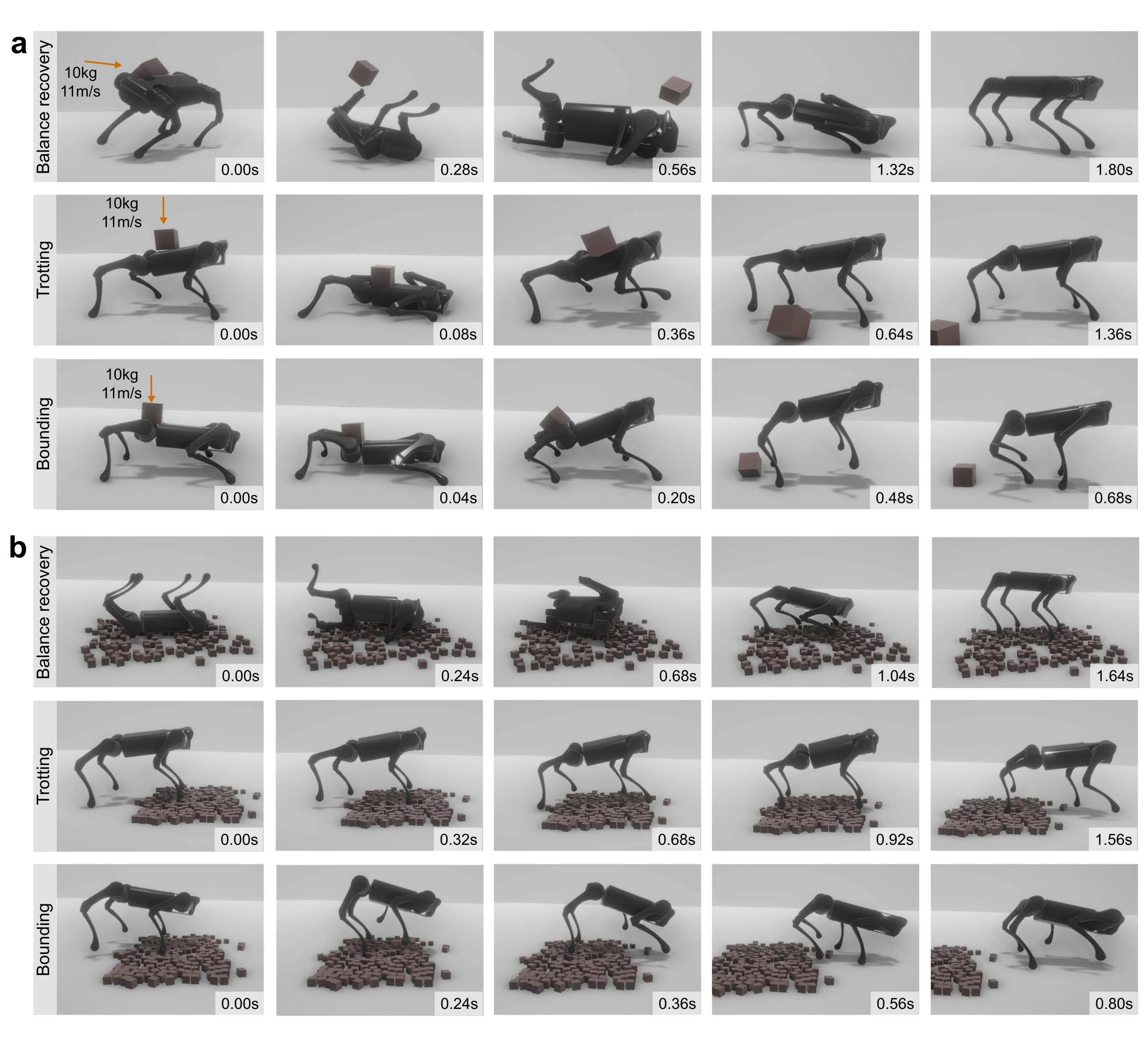}}
    \caption[Robustness tests of full-state policies for balance recovery, trotting and bounding against unexpected perturbations.]{Robustness tests of full-state policies for balance recovery, trotting and bounding against unexpected perturbations. \figcap{A} Locomotion under full-state policies perturbed by throwing a $10 kg$ box at the robot from a distance of $1 m$ with an initial velocity of $11 m/s$ from the right side of the robot body for balance recovery (top), and from the top for trotting (middle) and bounding (bottom). \figcap{B} Locomotion under full-state policies over unseen rubble for balance recovery (top), trotting (middle) and bounding (bottom).}
    \label{robust-full}    
\end{figure}

\begin{figure}[H]    
    \centering
        {\includegraphics[width=0.9\textwidth]{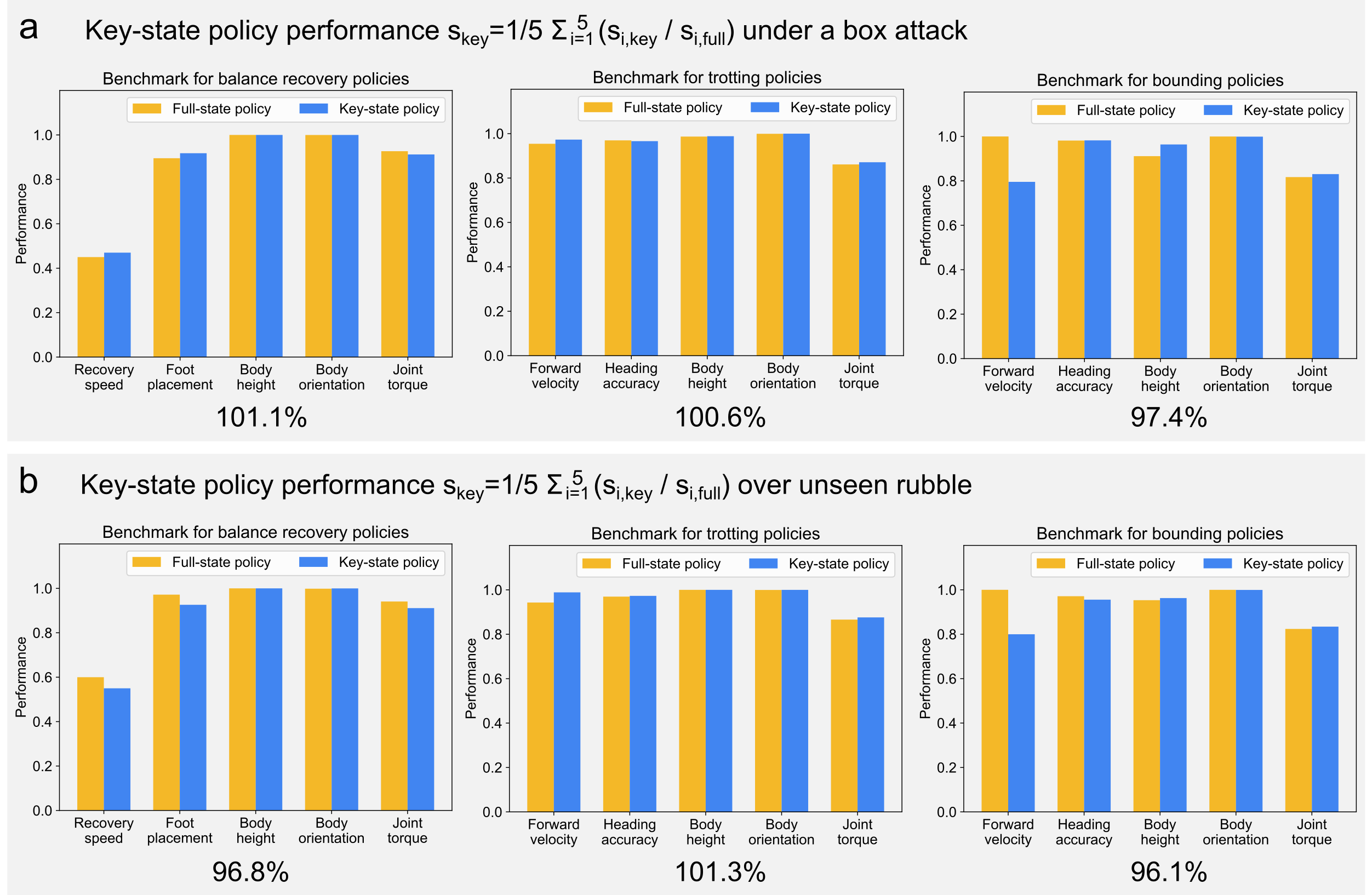}}
    \caption[Comparison of performance metrics for full-state policies and key-state policies for balance recovery, trotting, and bounding in two unseen scenarios.]{Comparison of performance metrics for full-state policies and key-state policies for balance recovery, trotting, and bounding in two unseen scenarios (corresponding to snapshots in Extended Data Figure \ref{ex:robust-key} and Supplementary Figure \ref{robust-full}). \figcap{A} Performance metrics of full-state policies and key-state policies under the box attack for balance recovery (left), trotting (middle) and bounding (right). The overall performance (balance recovery, trotting, bounding) of key-state policies with respect to full-state policies is 101.1\%, 100.6\% and 97.4\%, respectively. \figcap{B} Performance metrics of full-state policies and key-state policies over unseen rubble for balance recovery (left), trotting (middle) and bounding (right). The overall performance of key-state policies with respect to full-state policies is 96.8\%, 101.3\% and 96.1\%, respectively.}
    \label{robust-metric}    
\end{figure}

\begin{figure}[H]    
    \centering
        {\includegraphics[width=0.9\textwidth]{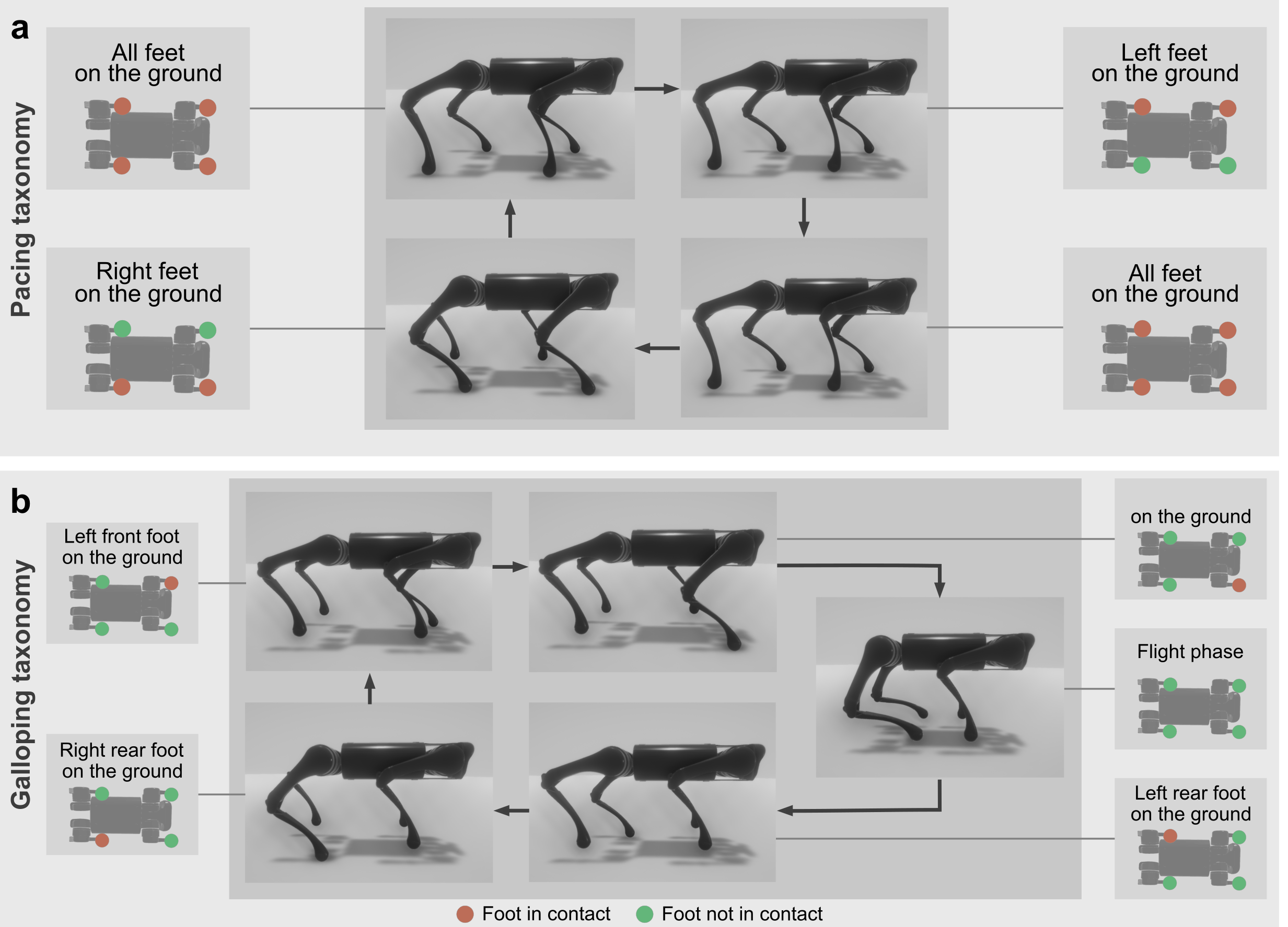}}
    \caption[Key-pose taxonomy for learning pacing and galloping gaits.]{Key-pose taxonomy for learning pacing and galloping gaits. \figcap{A} Key-pose taxonomy for pacing, including four key poses. \figcap{B} Key-pose taxonomy for galloping, including five key poses.}
    \label{taxonomy}    
\end{figure}

\begin{figure}[H]    
    \centering
        {\includegraphics[width=0.9\textwidth]{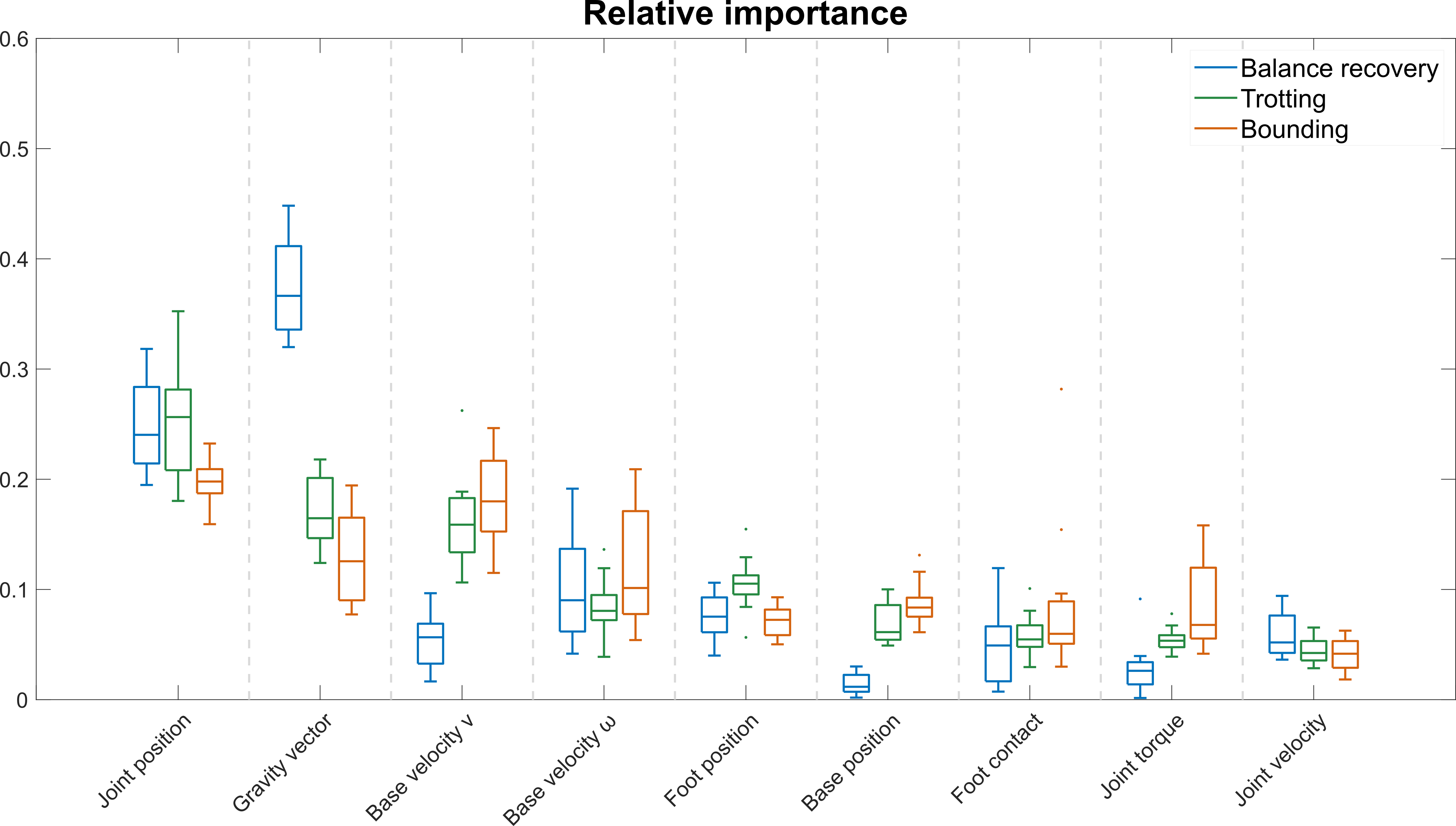}}
    \caption[A box plot showing statistics of relative importance of nine feedback states reflected by the maximum saliency value for balance recovery, trotting and bounding.]{A box plot showing statistics of relative importance of nine feedback states reflected by the maximum saliency value for balance recovery, trotting and bounding, based on the saliency maps in Supplementary Figure \ref{saliency-stand}-\ref{saliency-bound}.}
    \label{peak-box}    
\end{figure}

\begin{figure}[H]    
    \centering
        {\includegraphics[width=0.9\textwidth]{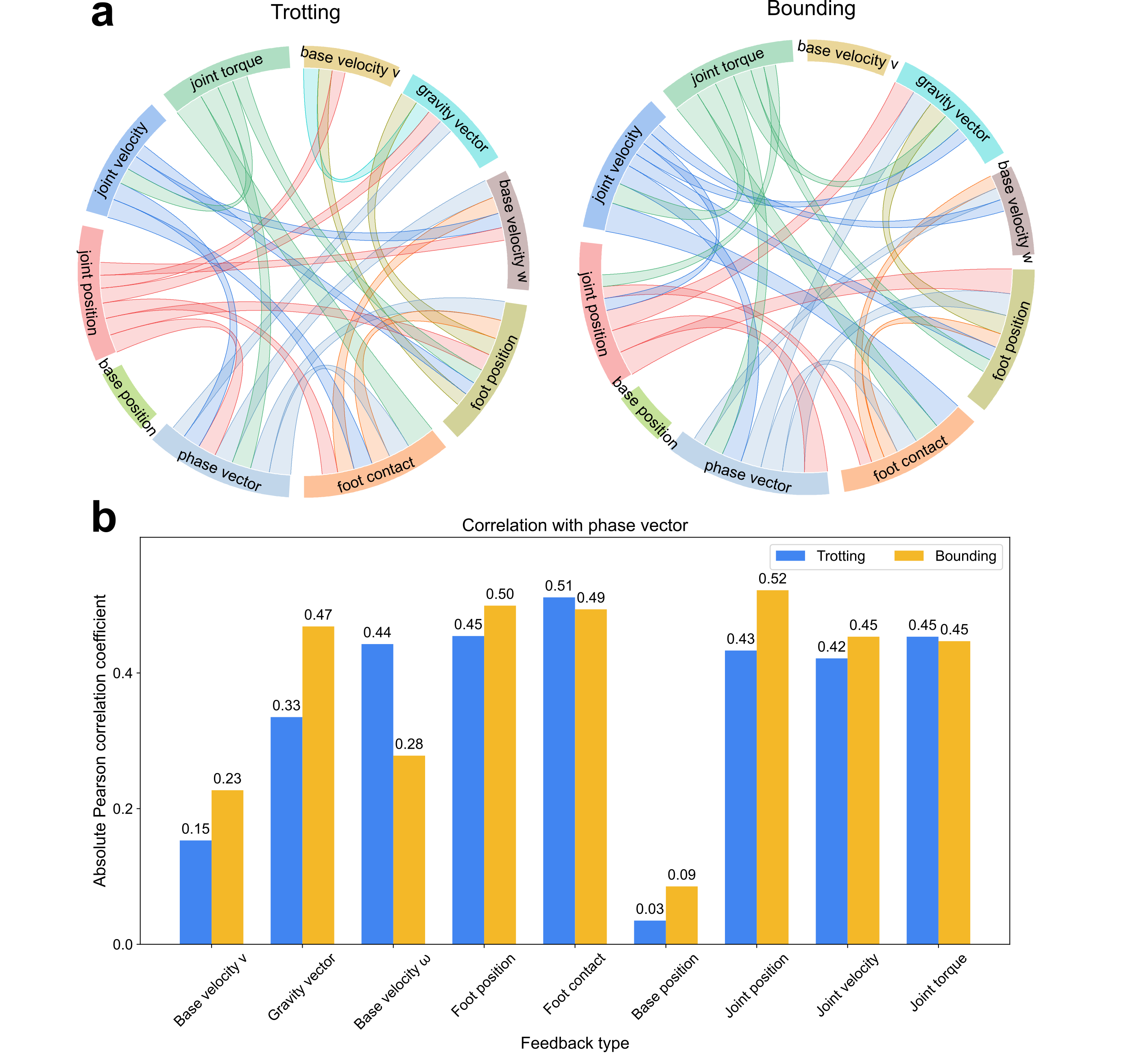}}
    \caption[Analysis of the correlation between the feedforward phase vector $(sin2\pi\phi, cos2\pi\phi)$ and nine feedback states for trotting and bounding.]{Analysis of the correlation between the feedforward phase vector $(sin2\pi\phi, cos2\pi\phi)$ and nine feedback states for trotting and bounding. \figcap{A} Chord diagrams showing the correlation between any two states (self-correlation and percentage $<50\%$ are removed for clarity), where the wider the link between the two states, the stronger they correlate with each other. \figcap{B} Bar plot showing the absolute Pearson correlation coefficient between the phase vector and each of the nine feedback states.}
    \label{phase-foot}    
\end{figure}

\pagebreak
\pdfbookmark[2]{Supplementary tables}{Supplementary tables}
\section*{Supplementary tables}
\captionsetup[table]{labelfont={bf},labelformat={default},labelsep=period,name={Supplementary Table},list=yes}

\begin{table}[H]
    \caption[List of full feedback states with dimensionality and the order of each dimension on saliency maps.]{List of full feedback states with dimensionality and the order of each dimension on saliency maps (FL: front left, FR: front right, RL: rear left, RR: rear right).}
    \label{tab:dimension}
    \centering
    \begin{tabular}{p{5cm}c}
        \hline
        Feedback state (Dimension)& Saliency map order  \\
        \hline
        Joint position (12)& FL, FR, RL, RR (Hip roll, hip pitch, knee)\\
        Gravity vector (3)& $\phi_x$, $\phi_y$, $\phi_z$\\
        Base linear velocity (3)& $v_x$, $v_y$, $v_z$\\
        Base angular velocity (3)& $\omega_x$, $\omega_y$, $\omega_z$\\
        Foot position (12)& FL, FR, RL, RR ($f_x$, $f_y$, $f_z$)\\
        Base position (3)& $x$, $y$, $z$\\
        Foot contact (4)& FL, FR, RL, RR\\
        Joint torque (12)& FL, FR, RL, RR (Hip roll, hip pitch, knee) \\
        Joint velocity (12)&FL, FR, RL, RR (Hip roll, hip pitch, knee) \\
        \hline
    \end{tabular}
\end{table}

\begin{table}[H]
    \caption[Standard deviation for Gaussian noise in feedback states for robustness tests.]{Standard deviation for Gaussian noise in feedback states for robustness tests. We adopt the noise level which is obtained from real robot measurements \cite{rudin2021learning}, and double the noise level as standard deviation of the Gaussian noise. The noise sampled from the corresponding Gaussian distribution with zero mean is added to each feedback state.}
    \label{tab:noise}
    \centering
    \begin{tabular}{p{5cm}m{3.5cm}}
        \hline
        Feedback state & Standard deviation  \\
        \hline
        Joint position & 0.02 $rad$ \\
        Gravity vector & 0.01 \\
        Base linear velocity & 0.02 $m/s$ \\
        Base angular velocity & 0.4 $rad/s$ \\
        \hline
    \end{tabular}
\end{table}

\begin{table}[H]
    \caption{Joint-level PD control gains for \textit{A1} quadruped robot.}
    \label{tab:pd}
    \centering
    \begin{tabular}{p{3cm}m{2cm}m{2cm}m{2cm}}
        \hline
         & Hip roll & Hip pitch & Knee pitch  \\
        \hline
        $K_p$ $(Nm/rad)$ & 100 & 100 & 100\\
        $K_d$ $(Nms/rad)$ & 5 & 5 & 5 \\
        \hline
    \end{tabular}
\end{table}

\begin{table}[H]
	\centering
	\caption{Specification of the \textit{A1} quadruped robot.}
	\label{tab:robot}
	\begin{tabular}{ l llcc }
	    \hline
         & & \hspace{-11mm} Joint range $(^{\circ})$ & $\tau$ $(Nm)$ & $\omega$ $(rad/s)$ [\textit{Peak}] \\
	    \cline{1-5}
	    & Hip roll & \hspace{-9mm} -46 -- 46 & 33.5 & 12 [\textit{21}] \\
	    & Hip pitch & \hspace{-10.5mm} -240 -- 60 & 33.5 & 12 [\textit{21}] \\
        & Knee & \hspace{-9mm}52.5 -- 154.5 & 33.5 & 12 [\textit{21}] \\
	    \cline{1-5}
		& Body size (LWH) & \multicolumn{3}{c}{0.361$m
		\times$0.194$m\times$0.114$m$} \\
		& Leg length (upper, lower) & \multicolumn{3}{c}{0.2$m$, 0.2$m$} \\
		\hline
	\end{tabular}
\end{table}

\begin{table}[H]
	\centering
	\caption{Mathematical formulation of the reward terms for quadruped locomotion.}
	\label{tab:reward}
	\begin{tabular}{ p{6cm}p{9cm}}
		\hline
		Physical quantities& Reward term \\ \hline
		Base orientation& $\varphi(\phi, [0,0,-1], \alpha), \quad\quad\alpha=-2.35$ \\ 
		Base height& $\varphi(h, \hat{h}, \alpha), \quad\quad\alpha=-51.16$ \\ 
		Base linear velocity& $\varphi(v_{base}^{world}, \hat{v}_{base}^{world}, \alpha), \quad\quad\alpha=-18.42$ \\ 
		Joint torque regularisation& $\varphi(\tau, 0, \alpha), \quad\quad\alpha=-0.004$          \\ 
		Joint velocity regularisation& $\varphi(\dot{q}, 0, \alpha), \quad\quad\alpha=-0.032$ \\ 
		Body ground contact& $
		\begin{cases}
		0, &\text{main body in contact with ground}\\
		1, &\text{main body not in contact with ground}
		\end{cases}
		$ \\ 
		Foot ground contact& $
		\begin{cases}
		0, &\text{no foot in contact with ground}\\
		1, &\text{foot in contact with ground}
		\end{cases}
		$ \\ 
		Symmetric foot placement (recovery)&
        $\varphi(p_{foot}^{base}, \hat{p}_{foot}^{base}, \alpha), \quad\quad\alpha=-51.16$ \\
		Symmetric foot placement (gaits)&
        $\varphi(1/4 \sum_{n=1}^{4}(p_{foot, n}^{world}), p_{base}^{world}, \alpha), \quad\quad\alpha=-51.16$\\
		Swing and stance &
		$\varphi(h_{foot}^{world}v_{foot}^{world},\hat{h}_{foot}^{world}v_{foot}^{world}, \alpha), \quad\quad\alpha=-460.50$ \\
		Reference foot contact& $
		\begin{cases}
		0, &\text{not match desired foot contact}\\
		1, &\text{match desired foot contact}
		\end{cases}
		$ \\ 
	    Yaw velocity& $\varphi(\dot{\psi}, 0, \alpha), \quad\quad\alpha=-7.47$          \\ 
		\hline
	\end{tabular}
\end{table}

\begin{table}[H]
	\centering
	\caption{Weights for the reward terms for different locomotion tasks.}
	\label{tab:weight}
	\begin{tabular}{ p{5cm}ccccc}
		\hline
		Physical quantities& Recovery & Trotting & Bounding & Pacing & Galloping \\ \hline
		Base orientation& 0.189 & 0.068 & 0.068 & 0.068 & 0.068\\ 
		Base height& 0.189 & 0.068 & 0.068 & 0.068 & 0.068\\ 
		Base linear velocity& 0.114 & 0.170 & 0.170 & 0.170 & 0.170\\ 
		Joint torque regularisation& 0.076 & 0.017 & 0.017 & 0.017 & 0.017\\ 
		Joint velocity regularisation& 0.076 & 0.017 & 0.017 & 0.017 & 0.017\\ 
		Body ground contact& 0.083 & 0.048 & 0.048 & 0.048 & 0.048\\ 
		Foot ground contact& 0.083 & 0.000 & 0.000 & 0.000 & 0.000\\ 
		Symmetric foot placement & 0.189 & 0.034 & 0.034 & 0.034 & 0.034\\ 
		Swing and stance &0.000 & 0.034 & 0.034 & 0.034 & 0.034\\
		Reference foot contact& 0.000 & 0.476 & 0.476 & 0.476 & 0.476\\ 
		Yaw velocity& 0.000 & 0.068 & 0.068 & 0.068 & 0.068\\ 
        \hline
	\end{tabular}
\end{table}

\begin{table}[H]
    \caption{Hyperparameters for SAC.}
    \label{tab:sac}
    \centering
    \begin{tabular}{p{5cm}m{3cm}}
        \hline
        Hyperparameter & Value  \\
        \hline
        Learning rate & 3e-4 \\
        Weight decay & 1e-6 \\
         \multirow{2}{*}{Discount factor}&0.995 (recovery)\\&0.955 (gaits) \\
        Soft target update & 0.001 \\
        Replay buffer size & 1e6 \\
        Steps per epoch & 5e3 \\
        \hline
    \end{tabular}
\end{table}

\pagebreak
\pdfbookmark[2]{Supplementary notes}{Supplementary notes}
\section*{Supplementary notes}

\pdfbookmark[3]{Deep reinforcement learning framework}{Deep reinforcement learning framework}
\subsection*{Deep reinforcement learning framework}
The objective of DRL is to learn a policy which can generate optimal actions given states observed from the environment. We used the Soft Actor Critic (SAC) algorithm \cite{haarnoja2018soft} incorporated with smoothing loss to train the policy \cite{yang2020multi}. Here we provide details about essential components of our DRL framework for learning locomotion skills, including state observation, action space, neural network architecture, reward terms, control framework and training procedure.

\pdfbookmark[4]{State observation and action space}{State observation and action space}
\subsubsection*{State observation and action space}
The design of state observation is detailed in Methods. Regarding the action space, we adopted the design from previous work \cite{yang2020multi}, since it has been shown to improve performance for certain motor tasks \cite{peng2017learning}. Specifically, the actions are twelve desired joint positions for a quadruped robot. 

\pdfbookmark[4]{Neural network architecture}{Neural network architecture}
\subsubsection*{Neural network architecture}
\textit{\textbf{Policy network.}}
A stochastic policy $\pi_{\theta}(s_t)$ is used as the actor in SAC for training, which is modeled as a Gaussian distribution $\mathcal{N} (\mu_{\theta} (s_t), \sigma_{\theta} (s_t)^2)$ with mean $\mu_{\theta} (s_t)$ and standard deviation $\sigma_{\theta} (s_t)$, where $\theta$ is the parameter set of a fully-connected neural network and $s_t$ is the state observed from the environment at time step $t$. The policy network has two hidden layers, each with 256 neurons and using a ReLU activation function. Suppose we have network input $\mathbf{X}\in\mathbb{R}^x$, the network output $\mathbf{Y}\in\mathbb{R}^{24}$ is computed as follows:

\begin{equation}
 \mathbf{y_1} = \mathrm{ReLU}(\mathbf{w}_0 \mathbf{X} + \mathbf{b}_0 )
\end{equation}

\begin{equation}
 \mathbf{y_2} = \mathrm{ReLU}(\mathbf{w}_1 \mathbf{y_1} + \mathbf{b}_1 ) 
\end{equation}

\begin{equation}
 \mathbf{Y} = \mathrm{tanh}(\mathbf{w}_2 \mathbf{y_2} + \mathbf{b}_2 )   
\end{equation}

where, $\mathbf{w}_0 \in \mathbb{R}^{256\times x},\mathbf{w}_1\in \mathbb{R}^{256\times256},\mathbf{w}_2\in \mathbb{R}^{24\times 256}$ are neural network weights, $\mathbf{b}_0\in \mathbb{R}^{256},\mathbf{b}_1\in \mathbb{R}^{256},\mathbf{b}_2\in \mathbb{R}^{24}$ are neural network biases. The network output $\mathbf{Y}$ is in the range of $[-1,1]$, which is then scaled by the corresponding joint range (see Supplementary Table \ref{tab:robot}). When deploying the learned policy on the robot, we use the deterministic mean outputs of the stochastic policy $\mu_{\theta}(s_t)$ as the references of joint angles. 

For full-state policies, the network input dimension is $x=64$ for balance recovery, and $x=66$ during trotting and bounding. For key-state policies, the network input dimension is $x=15, x=20$ for balance recovery and trotting, respectively, and $x=23$ for bounding, pacing and galloping.

\textit{\textbf{Critic network.}}
The critic in SAC consists of two Q functions which are encoded as two separate fully-connected neural networks. Similar to policy network, each critic network has two hidden layers, each with 256 neurons and using a ReLU activation function. The network input includes state and action, and the output is a scalar value.

\pdfbookmark[4]{Reward terms}{Reward terms}
\subsubsection*{Reward terms}
The reward function designed for learning quadruped locomotion is composed of eleven weighted reward terms related to the following physical quantities \cite{yang2020multi}: (i) base orientation $\phi$, (ii) base height $h$, (iii) base linear velocity $v$, (iv) joint torque $\tau$, (v) joint velocity $\dot{q}$, (vi) body ground contact $C_{bg}$, (vii) foot ground contact $C_{fg}$, (viii) symmetric foot placement $p_{foot}$, (ix) swing and stance $h_{foot}v_{foot}$, (x) reference foot contact $\hat{C}_{fg}$, and (xi) yaw velocity $\dot{\psi}$. Balance recovery uses reward terms (i)-(viii), trotting, bounding, pacing and galloping use reward terms (i)-(xi). We use a radial basis function (RBF) to formulate each reward term related to continuous physical quantities:

\begin{equation}
\varphi(x, \widehat{x}, \alpha)=\exp \left(\alpha(\hat{x}-x)^{2}\right)    
\end{equation}

where $x$ is the continuous physical quantity, $\hat{x}$ is the corresponding reference, and $\alpha$ is the shape parameter which controls the width of RBF. Reward terms (vi), (vii) and (x) which are related to discrete physical quantities are designed separately. The detailed mathematical formulation and weight for each reward term for the five locomotion tasks are in Supplementary Table \ref{tab:reward} and \ref{tab:weight}, respectively.

\pdfbookmark[4]{Control framework}{Control framework}
\subsubsection*{Control framework}
Our control framework for legged locomotion consists of a high-level behavior loop running the policy network at $25 Hz$ and a low-level joint position tracking loop running joint-level PD controllers at $1 kHz$. In the high-level control loop, the measured or estimated states are first passed to a first-order low-pass butterworth filter with a cut-off frequency of $10 Hz$. The policy network receives the filtered states as input and generates desired joint positions at $25 Hz$. In the low-level control loop, the desired joint positions from the neural network output are filtered by a first order low-pass butterworth filter with a cut-off frequency of $4 Hz$. Joint-level PD controllers receive the filtered joint position references and the measured joint positions, and generate joint torques at $1 kHz$ according to the following equation:

\begin{equation}
\tau=K_p(q_{d}-q_{m})-K_d\dot{q}_{m}
\end{equation}

where, $\tau$ is the computed joint torque, $q_{d}$ is the desired joint position, $q_{m}$ is the measured joint position, and $K_p, K_d$ are PD gains (see Supplementary Table \ref{tab:pd}).

\pdfbookmark[4]{Training procedure}{Training procedure}
\subsubsection*{Training procedure}
\textit{\textbf{Soft Actor Critic with smoothing loss.}}
Soft Actor Critic learns an optimal policy $\pi^*(s_t)$ by maximizing the expected sum of rewards and entropy: 

\begin{equation}
    J_{SAC} (\pi(s_t)) = \sum_{t=0}^{T} \mathbb{E}_
{(s_t ,a_t) \sim \rho_{\pi}} [r(s_t, a_t)
+ \alpha H(\pi(\cdot| s_t))]
\end{equation}

where $s_t$ and $a_t$ are the observed state and action at time step $t$, $\rho_{\pi}$ is the sample distribution, $r(s_t, a_t)$ is the reward received at time step $t$, $H(\pi(\cdot| s_t))$ is the entropy of policy $\pi$ at time step $t$, and $\alpha$ is temperature parameter which is automatically tuned in SAC to balance exploitation and exploration. To minimize the applied joint torques and generate smooth robot motions, we introduce an additional smoothing loss term $J_{smooth}$ as in our previous work \cite{yang2020multi}:

\begin{equation}
    J_{smooth}(\mu (s_t))=||\mu(s_t)-q_m||_2
\end{equation}

where $\mu(s_t)$ are the deterministic mean outputs of the stochastic policy used as the desired joint references, and $q_m$ are the measured joint positions. Thus the goal of the DRL agent in this work is to find a policy that maximizes the following objective:

\begin{equation}
    J(\pi(s_t))=J_{SAC} (\pi(s_t))-J_{smooth}(\mu (s_t))
\end{equation}

Meanwhile, SAC also adopts Double Q-learning to learn two Q functions, details of which can be found in the literature \cite{hasselt2010double,van2016deep}. The SAC hyperparameters are shown in Supplementary Table \ref{tab:sac}.

\textit{\textbf{Initialization.}}
In this work, we used two methods to initialize the robot pose during training for comparison, i.e., random-pose initialization and key-pose initialization. The pose of the robot consists of: base height, base orientation (roll and pitch angles) and twelve joint positions. We note that base linear velocity and base angular velocity were both set as zero at the start of each episode. Key-pose initialization is introduced in Methods. For random-pose initialization, the robot was initially at a fixed height of $0.4 m$, yaw angle was set as zero, roll and pitch angles were both sampled from a uniform distribution $[-180^{\circ},180^{\circ}]$, and each of the twelve joint positions was sampled from a uniform distribution of its corresponding joint range $[q_{min},q_{max}]$. For balance recovery, the distribution of joint positions was the same as the joint range of the \textit{A1} quadruped robot, seen in Supplementary Table \ref{tab:robot}. For trotting and bounding, a narrower distribution of joint position was used for a more efficient exploration \cite{reda2020learning}, i.e., $[-30^{\circ},30^{\circ}]$ for hip roll angle, $[-60^{\circ},60^{\circ}]$ for hip pitch angle, and $[52.5^{\circ},120^{\circ}]$ for knee angle.

\textit{\textbf{Early termination.}}
During training, each episode can proceed either for a fixed time period or until certain termination criteria are met. For balance recovery, each episode terminates after a fixed time period of $10 s$. For trotting, bounding, pacing and galloping, we designed early termination criteria in order to augment the samples collected by the agent. Specifically, an episode will terminate if any body segment excluding feet is in contact with the ground or body orientation exceeds $90^\circ$. Otherwise, the episode terminates after $10 s$.

\pdfbookmark[3]{Performance benchmark}{Performance benchmark}
\subsection*{Performance benchmark}
As shown by the time-elapsed snapshots of balance recovery, trotting and bounding in Fig. \ref{fig:benchmark}\capA-\capC, the key-state policy can successfully acquire all the skills equally well as the full-state policy. Using only the key states, the robot was fully able to recover from various fallen poses to a standing posture, such as lying down on its back (Fig. \ref{fig:benchmark}\capA) and many other configurations (Supplementary Figure \ref{recovery-snapshots}). Moreover, it can perform trotting (Fig. \ref{fig:benchmark}\capB) and bounding gaits (Fig. \ref{fig:benchmark}\capC) successfully and stably. However, irrelevant-state policies fail to move forward for trotting and bounding (see Supplementary Video 1).

The bar plots in Fig. \ref{fig:benchmark}\capD~show the values of performance metrics of all the task-related physical quantities for the full-state policy and the key-state policy, which are nearly identical to each other. There are three nuances: balance recovery on the recovery speed and final foot placement; bounding on the forward velocity. Although these quantities slightly differ in terms of numerical values, both policies perform well and have little differences in terms of the actual task completion.

For the balance recovery speed, such as where the robot stands up from lying on its back in Fig. \ref{fig:benchmark}\capA, it takes $1.48 s$ for the full-state policy and $1.84 s$ for the key-state policy to complete. Regarding the final foot placement, as seen in the robot snapshots in Fig. \ref{fig:benchmark}\capA, both policies control the robot to stand with four feet placed symmetrically at the end of recovery. As for the subtle difference in the forward bounding velocity, we observed a slightly lower average steady-state velocity which comes from a naturally learned stable gait, as there was no penalty if the measured forward velocity is higher than the desired velocity specified during training (see Methods). More detailed data plots of task-related physical quantities can be found in Supplementary Figure \ref{benchmark-detail}, for the full-state and key-state policies of three locomotion tasks.

\pdfbookmark[3]{Robustness tests}{Robustness tests}
\subsection*{Robustness tests}\label{sec:robust}
We validated the robustness of key-state policies in two categories of uncertainties: (i) environmental uncertainties, i.e., scenarios that were not encountered during training; and (ii) robot uncertainties, i.e., sensing noises, and variations of robot mass and control gains. All the robust locomotion tests on the \textit{A1} robot in these unseen scenarios can be found in Supplementary Video 3.

We designed two unseen scenarios reflecting environmental uncertainties: (i) locomotion under box attack (Extended Data Figure \ref{ex:robust-key}\capA) and (ii) locomotion over uneven rubble (Extended Data Figure \ref{ex:robust-key}\capB), which were applied to balance recovery, trotting and bounding. In the box-attack scenario, a $10 kg$ box was thrown at the robot's body from $1 m$ away at an initial velocity of $11 m/s$, from the right side of the robot (for balance recovery) and from the top (for trotting and bounding). The robot was able to withstand all these force perturbations and continue the task when using the key-state policies (Extended Data Figure \ref{ex:robust-key}\capA). In the uneven rubble scenario, 200 cubes ($3 cm$ edge length) were randomly piled within an $0.8m\times0.8m$ area to test balance recovery policies; for trotting and bounding, 200 cubes were randomly distributed in an area of $0.8m\times0.5m$. The robot was able to succeed in all the rubble scenarios across three locomotion tasks using key-state policies (Extended Data Figure \ref{ex:robust-key}\capB).

Besides, we also tested the robustness of full-state policies in the same scenarios as a baseline for comparison (Supplementary Figure \ref{robust-full}). We concluded that key-state policies can achieve task performance equally well in all three locomotion tasks, compared to that of the full-state policies, with an average performance of 99.7\% in the box-attack scenario and 98.1\% in the uneven rubble scenario (see quantified metrics in Supplementary Figure \ref{robust-metric}). 

Moreover, the key-state policies also demonstrated robustness against robot uncertainties in the following aspects:
(i) noisy feedback states, where the noise was sampled from a Gaussian distribution with zero mean and corresponding standard deviation (see details in Supplementary Table \ref{tab:noise}) and added to the corresponding feedback state;
(ii) robot mass variation, where the original robot mass was changed by $\pm10\%$;
and (iii) variation of the control gain, i.e., the proportional gain $K_p$ for joint-level PD control was altered by $\pm10\%$ away from the original $K_p$ value (Supplementary Table \ref{tab:pd}). The \textit{A1} robot was able to successfully complete all the robustness tests, and see Supplementary Video 3 for the robust locomotion skills in action.

\pdfbookmark[3]{Evaluation of impact of sensing accuracy}{Evaluation of impact of sensing accuracy}
\subsection*{Evaluation of impact of sensing accuracy}
We can study the impact of feedback accuracy by analyzing and composing two sensitivity matrices: (i) Sensitivity of action with respect to the state changes, i.e., saliency maps; (ii) Sensitivity matrix representing the sensor noise levels for each type of feedback states. We formulate the sensitivity matrix of the sensor noise level $M_{sensor}$ as follows:

\begin{equation}
 M_{sensor}=\frac{\sigma_{sensor}}{s_{max}-s_{min}},  
\end{equation}

where $\sigma_{sensor}$ is the standard deviation of the sensor noise level, $s_{max}$ and $s_{min}$ is the upper and lower range of the state. Then, the sensitivity matrix of the action with respect to the measurement’s quality can be evaluated as follows by composing two sensitivity matrices element-wise:

\begin{equation}
  M=M_{saliency} \circ M_{sensor}  
\end{equation}

\pdfbookmark[3]{Comparison of relative importance using maximum saliency value}{Comparison of relative importance using maximum saliency value}
\subsection*{Comparison of relative importance using maximum saliency values}
Due to the phase dependencies of saliency analysis discussed in Results, we  conduct further investigation to validate whether it is appropriate to compare the relative importance of states by using the mean of saliency values over all time steps. We compare the results using both the mean and the maximum of saliency values to quantify the relative importance.

For a certain type of feedback state $o \in \mathbb{R}^h$, importance $I_o$ can be computed in two ways:

(i) add up the saliency values over all time steps for each dimension of a type of feedback and average the importance over dimensions:

\begin{equation}
    I_o=\frac{1}{h} \sum_{i=1}^{h}\sum_{t=1}^{N} S(o_{i,t})
\end{equation}

(ii) find the maximum saliency value over all time steps across all the dimensions of a type of feedback:

\begin{equation}
    I_o=\max_{t\in[1,N],i\in[1,h]} S(o_{i,t})
\end{equation}

where $S(o_{i,t})$ is the saliency value for the $i$-th dimension of feedback state $o$ at time step $t$, and $N$ is the number of total time steps during the entire motion. 

This work considers nine feedback states in total. For feedback state $o$, relative importance $r_o$ is defined as follows:

\begin{equation}
   r_o=\frac{I_o}{\sum_{o=1}^9 I_o} 
\end{equation}

Figure \ref{fig:saliency}\capC~ and Supplementary Figure \ref{peak-box} show the rankings of relative importance among states according to mean and maximum saliency values, respectively, based on the same set of saliency maps in Supplementary Figure \ref{saliency-stand}-\ref{saliency-bound}. While using maximum saliency value to rank the relative importance, we found that the key important states remain the same, although there are only slight variations \textit{within} the key states, e.g., gravity vector becomes the highest ranking feedback state, replacing joint positions. Therefore, we confirm that using the mean of saliency values over all time is an effective way to identify the key states for locomotion skills.

\end{document}